%% file: main.tex
\documentclass[runningheads]{llncs}

\usepackage{eccv}

\usepackage{eccvabbrv}

\usepackage{graphicx}
\usepackage{booktabs}

\usepackage[accsupp]{axessibility}  %

\usepackage{hyperref}

\usepackage{orcidlink}

\input{preamble}
\begin{document}

\title{Unlearning Under Imbalance: Benchmarking Fairness in Multimodal LLM Unlearning} 

\titlerunning{Unlearning Under Imbalance}

\author{Lorenzo Orsingher\thanks{Work conducted while the author was affiliated with University of Trento and Fondazione Bruno Kessler, \texttt{lorenzo.orsingher@gmail.com}}\orcidlink{0009-0002-2625-8512} 
\and
Thomas {De Min}\inst{1}\orcidlink{0009-0008-4283-5555} \and
Massimiliano Mancini\inst{1}\orcidlink{0000-0001-8595-9955} \and
Davide Talon\inst{2}\orcidlink{0009-0003-6029-1532} \and
Elisa Ricci\inst{1, 2}\orcidlink{0000-0002-0228-1147}
}

\authorrunning{L.~Orsingher et al.}

\institute{University of Trento, Trento, Italy \\
\email{\{thomas.demin, massimiliano.mancini, e.ricci\}@unitn.it}
\and
Fondazione Bruno Kessler, Trento, Italy\\
\texttt{dtalon@fbk.eu} \\
\vspace{5pt}
\faHome~\href{https://lorenzoorsingher.github.io/unlearning-under-imbalance}{\textcolor{linkblue}{lorenzoorsingher/unlearning-under-imbalance}
}
}

\maketitle

\input{sec/0_abstract}    
\input{sec/1_intro}

\input{sec/2_related}
\input{sec/3_problem}

\input{sec/4_benchmark}

\input{sec/5_method}

\input{sec/6_experiments}
\input{sec/7_conclusion}

\section*{Acknowledgements}
We acknowledge the CINECA award under the ISCRA initiative for the availability of high-performance computing resources and support. This work is supported by the EU project and ELLIOT (101214398). 
This paper has been supported by the European Union’s Horizon Europe research and innovation actions under grant agreement No 101215032.
This work was partially funded by the European Commission’s Swarmchestrate Horizon Europe project (No. 101135012).
Thomas De Min is funded by NextGeneration EU.

\bibliographystyle{splncs04}
\bibliography{main}

\input{sec/A_data_generation}

\input{sec/E_metrics}
\input{sec/C_implementation_details}

\input{sec/D_additional_results}
\input{sec/F_naive_pca}
\input{sec/G_qualitatives}

\end{document}

%% file: preamble.tex
\usepackage{comment}
\usepackage{soul}
\usepackage{tabularx}
\usepackage{pifont}
\usepackage{algorithm}
\usepackage{algorithmic}
\usepackage{amsmath}
\usepackage{multirow}
\usepackage{makecell}
\usepackage{tikz}
\usepackage{wrapfig,lipsum,booktabs}
\usepackage{fontawesome5}

\usepackage{listings}
\usepackage[table]{xcolor}
\usepackage{mdframed}
\usepackage{colortbl}
\usepackage{arydshln}

\newcommand{\targetattr}[1]{\textcolor{OrangeRed}{#1}}
\newcommand{\protectedattr}[1]{\textcolor{Cerulean}{#1}}

\newcommand{\ourbenchmark}{FAIRGET}
\newcommand{\ourmethod}{FAUN}
\newcommand{\ourmethodlong}{\textbf{F}air \textbf{A}ctivation \textbf{UN}learning}

\renewcommand{\paragraph}[1]{\vspace{1pt}\noindent\textbf{#1}}
\newcommand{\smalltt}[1]{{\small\texttt{#1}}}

\newcommand{\checkm}{{\color{ForestGreen}\ding{51}}}
\newcommand{\crossm}{{\color{red}\ding{55}}}

\newcommand{\avggap}{avg.\ gap}

\newcommand{\layer}{\ell}
\newcommand{\weights}[1][{}]{\theta_\mathit{#1}} 
\newcommand{\model}[1][{}]{\pi_{\weights[#1]}}
\newcommand{\dataset}[1][{}]{\mathcal{D}_\mathit{#1}}
\newcommand{\sample}[1][{}]{x_\mathit{#1}}
\newcommand{\target}[1][{}]{y_\mathit{#1}}
\newcommand{\rephrasingf}[1][{}]{\psi(#1)}
\newcommand{\pred}[1][{}]{\hat{y}_\mathit{#1}}
\newcommand{\answer}[1][{}]{o_\mathit{#1}}
\newcommand{\predanswer}[1][{}]{\hat{o}_\mathit{#1}}
\newcommand{\attribute}[1][{}]{a_\mathit{#1}}

\newcommand{\targetspace}{\mathcal{Y}}
\newcommand{\attributespace}{\mathcal{A}}

\newcommand{\groupspace}{\mathcal{G}}

\newcommand{\bsample}[1][{}]{b_\mathit{#1}}
\newcommand{\bdata}[1][{}]{\mathcal{B}_\mathit{#1}}
\newcommand{\bcardinality}[1][{}]{M_\mathit{#1}}

\newcommand{\hidden}[1][{}]{h^\layer_\mathit{#1}}
\newcommand{\avghidden}[1][{}]{\bar{h}^\layer_\mathit{#1}}
\newcommand{\unbiasedhidden}[1][{}]{\tilde{h}^\layer_\mathit{#1}}
\newcommand{\allhidden}[1][{}]{H^\layer_\mathit{#1}}
\newcommand{\covariance}[1][{}]{\Sigma^\ell_\mathit{#1}}
\newcommand{\eigenvalues}{\Lambda}
\newcommand{\steering}[1][{}]{u^\layer_\mathit{#1}}
\newcommand{\unbiasedsteering}[1][{}]{\tilde{u}^\layer_\mathit{#1}}
\newcommand{\coeff}{\alpha}
\newcommand{\population}[1][{}]{N_\mathit{#1}}
\newcommand{\components}[1][{}]{V_\mathit{tr}^\mathit{#1}}
\newcommand{\topcomponents}{c}
\newcommand{\range}[2]{#1,\dots, #2}
\newcommand{\qwen}{Qwen2.5-VL-7B\xspace}

\newcommand{\idefics}{Idefics3-8B\xspace}

\newcolumntype{Y}{>{\centering\arraybackslash}X} %
\newcolumntype{C}[1]{>{\centering\arraybackslash}m{#1}} %
\newcolumntype{L}[1]{>{\raggedright\arraybackslash}m{#1}} %
\newcolumntype{R}[1]{>{\raggedleft\arraybackslash}m{#1}} %

\definecolor{experiment}{HTML}{F3F3F3}

\usepackage{xcolor}
\usepackage{listings}

\definecolor{codecomments}{HTML}{204680}
\definecolor{codegray}{rgb}{0.5,0.5,0.5}
\definecolor{codepurple}{rgb}{0.58,0,0.82}
\definecolor{codedef}{HTML}{32a891}
\definecolor{codefunction}{HTML}{298b96}
\definecolor{backcolour}{rgb}{0.98, 0.98, 0.98}

\definecolor{refs-blue}{HTML}{367dbd}

\definecolor{linkblue}{RGB}{45,105,180}

\usepackage{microtype}

\definecolor{oraclered}{HTML}{ad3232}

%% file: sec/0_abstract.tex
\begin{abstract}
Machine unlearning has emerged as a tool for removing personal data from trained models to comply with recent AI regulations.
To evaluate unlearning effectiveness in multimodal large language models (MLLMs), prior works fine-tune models on fictitious identities, simulating unlearning requests on subsets of these IDs, which are typically uniformly distributed. 
However, in realistic scenarios, people from different demographic groups may request to be unlearned at different frequencies, potentially altering the model’s internal beliefs for these groups and leading to biased behaviors.
To fill this gap, we propose FAIRGET, the first Visual Question Answering benchmark that evaluates unlearning under unbalanced, realistic, forget requests.
These requests are designed to simulate multiple realistic scenarios, ranging from simple to challenging settings, that lead to biased unlearned models if fairness is not accounted for.
Additionally, we propose FAUN, the first unlearning algorithm for MLLMs that forgets unlearning data while preserving model fairness. 
FAUN exploits a bias-aware activation steering mechanism to unlearn identities while accounting for the unbalanced nature of the forget data.
Experiments on FAIRGET and the established FIUBench demonstrate our method's superiority both in unlearning quality and fairness.
\keywords{Multimodal Large Language Models \and Unlearning \and Fairness}
\end{abstract}

%% file: sec/1_intro.tex
\section{Introduction}
\label{sec:intro}
Ensuring trustworthiness of multimodal large language models (MLLMs) and their compliance with emerging AI regulations~\cite{GDPR2016, CCPA2018, AIAct2024, WhiteHouseAIExecOrder2023} has become a critical challenge~\cite{zhang2023right, hamon2024three, judge2025code}. 
In particular, the \textit{right to be forgotten} (RTBF)~\cite{GDPR2016, CCPA2018} entitles individuals to request the removal of their personal information from trained models, thereby protecting their online privacy.
Yet, naively retraining MLLMs from scratch to remove one or more data instances can be extremely expensive and unfeasible for numerous requests.

\begin{figure}[t!]
     \begin{subfigure}[c]{0.48\textwidth}
        \centering
        \includegraphics[width=1\linewidth]{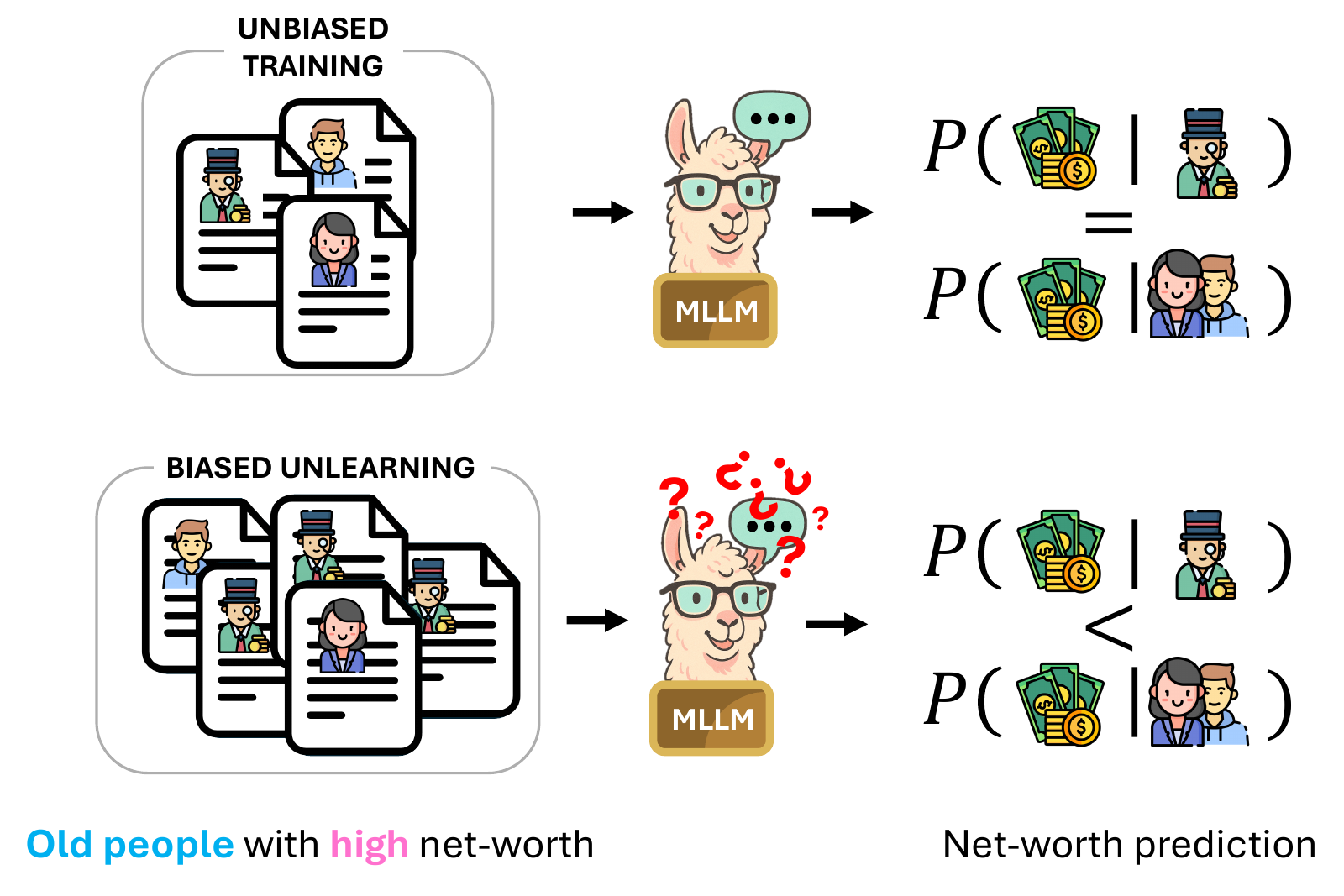}
    \end{subfigure}
    \hfill
    \begin{subfigure}[c]{0.48\textwidth}
        \centering
        \includegraphics[width=\linewidth]{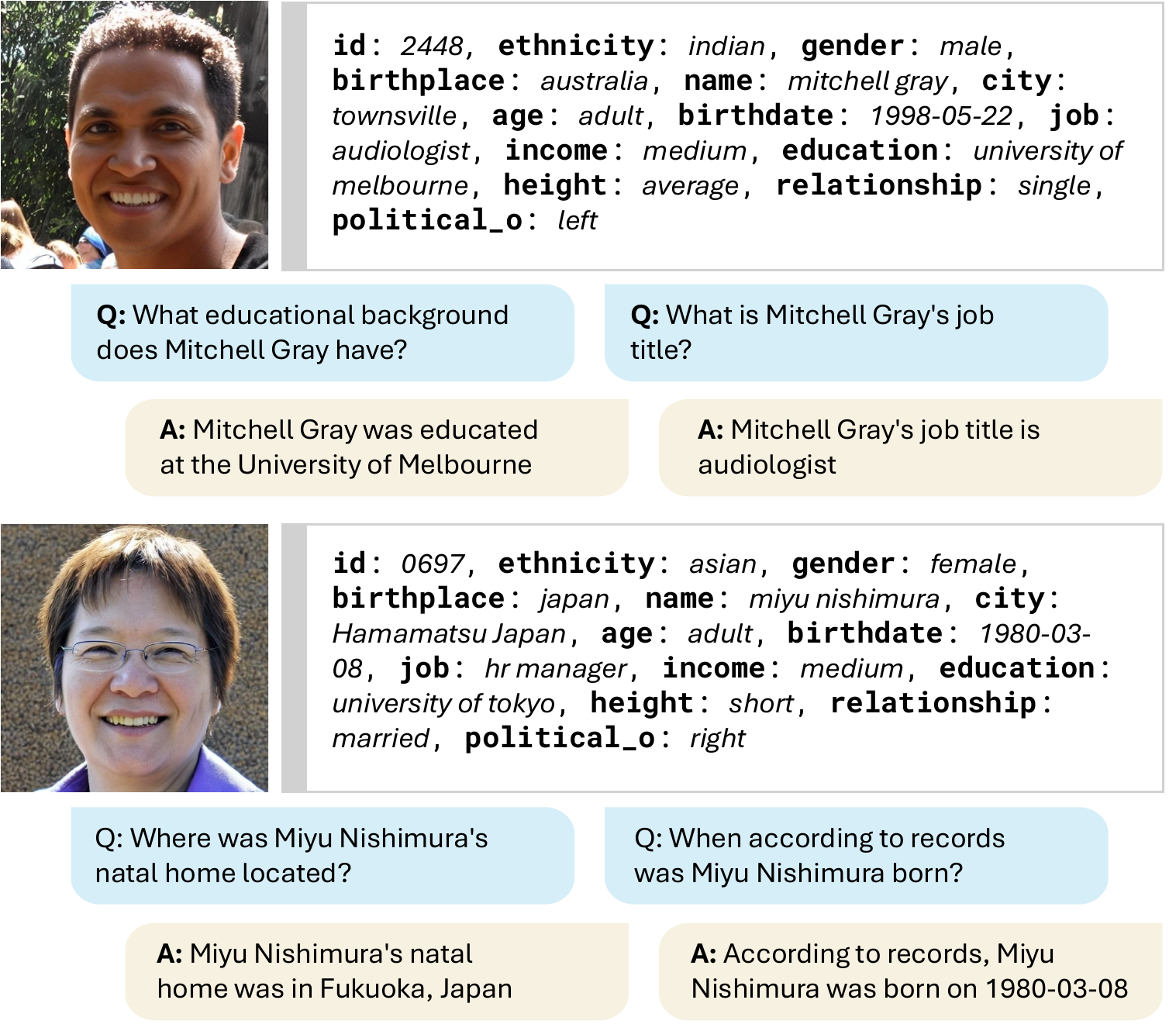}
    \end{subfigure}
    \caption{\textbf{Left:} We focus on MLLMs unlearning under unbalanced forget sets. If unlearning requests are unbalanced toward a specific subgroup (\eg, wealthy older individuals), the model fairness can degrade, causing \targetattr{target} attributes prediction to be biased by \protectedattr{protected} attributes.
    \textbf{Right:} By controlling the generation of visual and textual attributes, we can regulate the distribution of VQA pairs in \ourbenchmark{}, enabling fair pretraining while introducing intentionally unbalanced unlearning, which can result in unfair models.
 }
    \label{fig:teaser}
\end{figure}

To overcome this, approximate machine unlearning has emerged as a means to selectively remove specific data points or concepts from a trained model without requiring full retraining~\cite{bourtoule2021machine, cao2015towards, sekhari2021remember, ginart2019making}. %
In the RTBF case, unlearning aims to erase traces of information related to individuals that request to be forgotten, so that the model can no longer recall or rely on them for its predictions~\cite{mainitofu, dontsov-etal-2025-clear, liu2025protecting, mabenchmarking}. 
But, \textit{who asks to be forgotten?} 
Studies highlight that people value their privacy differently based on their economic status~\cite{epstein2020markers}, residence~\cite{cvrcek2006study}, or education~\cite{park2021data}: in \textit{real-world} settings, unlearning requests are inherently \textit{unbalanced} toward specific demographic groups~\cite{bertram2019five,de2025group}. 
As previous work highlight~\cite{de2025group}, the effects of unlearning unbalanced (\ie, non-\iid) data can extend beyond accuracy or privacy guarantees, as a model's internal belief about different groups can change, ultimately affecting its fairness.

Consider, for instance, a scenario where most RTBF requests originate from older individuals with high net-worth, like in \cref{fig:teaser}.
As the model forgets part of wealthy older individuals, it becomes biased and accounts for age when predicting people's net-worth, incorrectly associating older people with lower net-worth.
Multimodality further exacerbates the problem; omitting personal details in the textual modality at training time (\eg, ethnicity) does not prevent the model from memorizing them from visual inputs, learning spurious correlations.
While recent advances in machine unlearning for (multimodal) LLMs have explored selective forgetting of identities~\cite{fan2024simplicity,li2024single}, they overlook fairness concerns that arise from unbalanced unlearning requests.

To address this overlooked problem, we present \ourbenchmark{}, the first multimodal benchmark that evaluates unlearning robustness to unbalanced forget requests.
\ourbenchmark{} is composed of 4k fictitious identities, 40k images and 225k Q\&As about profiles, resulting in more than 10$\times$ the size of existing unlearning benchmarks (\cref{tab:benchmark,fig:teaser}). 
Our benchmark is obtained from a large pool of generated faces, curated to achieve balance among demographic groups (\ie,~age, gender, and ethnicity).
Q\&As are instead constructed from DeepSeek's~\cite{guo2025deepseek} generated templates, where attributes are filled via rule-based sampling, guaranteeing control over data generation (\eg, income, marital status, place of birth).
This structure allows us to obtain models that are fair when tuned on the dataset, but also to construct controlled and unbalanced unlearning requests.

We also present \ourmethodlong{} (\ourmethod{}), the first multimodal approach to jointly preserve privacy and model fairness.
Inspired by activation steering~\cite{turner2023steering,rimsky2024steering}, \ourmethod{} steers retain set activations with those of the forget set, computing gradient updates that make the model robust against forget set leaks.
Additionally, as forget set activations can be unbalanced towards a specific group, we introduce bias-informed Principal Component Analysis that suppresses dominant group components, producing ``bias-free'' forget set activations.

Extensive evaluations on \ourbenchmark{} show that \ourmethod{} achieves the best trade-off between unlearning indicators (41.97 \vs 42.31 forget EM, and 63.84 \vs 60.51 retain EM \wrt SimNPO) and fairness metrics (6.35 DP \vs 25.19 of SimNPO).
\ourmethod{} is also effective in FIUBench~\cite{mabenchmarking}, an established unlearning benchmark assuming balanced unlearning requests, while retaining general utility on MME~\cite{fumme}. {Unlike existing baselines, \ourmethod{} yields robust results across settings.}

\paragraph{Contributions:} (i) We introduce and formalize unbalanced multimodal identity unlearning, reflecting a more realistic and relevant scenario for current unlearning research; (ii) We present \ourbenchmark{}, a new {Visual Question Answering} (VQA) benchmark on fictitious identities %
 that systematically evaluate existing unlearning approaches under the novel unbalanced forget setting; (iii) We introduce \ourmethod{}, a novel method based on activation steering that permanently unlearns while preserving the model's fairness  and achieving SOTA results.

%% file: sec/2_related.tex
\section{Related Work}
\label{sec:related}

\paragraph{Unlearning.}
Early efforts in machine unlearning focus on certified data removal from small classifiers~\cite{bourtoule2021machine,yan2022arcane,aldaghri2021coded} by retraining (part of) the model to unlearn.
Yet, with the advent of (multimodal) LLMs, retraining to satisfy unlearning requests is computationally prohibitive~\cite{chundawat2023can,kurmanji2023towards,nguyen2025survey}.
Approximate unlearning overcome the need for model retraining by relaxing the certified removal constraint~\cite{kurmanji2023towards,jia2023model,fan2023salun}.
Examples of approximate unlearning methods include gradient ascent~\cite{yao2024large, mainitofu, chen2023unlearn, kurmanji2023towards} and random labeling~\cite{fan2023salun, yao2024large}, applied to the data to be forgotten.  
Although multimodal unlearning remains largely unexplored, recent advances have primarily focused on LLMs~\cite{shen2025lunar, fan2024simplicity, wangrethinking, li2024wmdp}.
Most recently, SimNPO~\cite{fan2024simplicity} refines NPO~\cite{zhangnegative} through a length-normalized preference mechanism that dynamically targets unlearning based on sample memorization, improving efficiency and achieving SOTA results in LLM unlearning.
Shen et al.\cite{shen2025lunar}, instead, propose the first activation steering approach for unlearning, though no extension has been proposed for multimodal. Finally, in the multimodal setting FTTP~\cite{li2025forget} introduces selective ascent steps only on those tokens related to personal information while SIU~\cite{li2024single} and PD~\cite{chen2025safeeraser} require tailored supervision (multifaced and Prompt Decouple, respectively).
To our knowledge, %
MIU~\cite{de2025group} is the only method tackling unlearning under unbalanced requests. It does so by minimizing the mutual information between image features and group information to address unbalanced unlearning requests.
However, it focused on image classification rather than MLLMs, where unlearning requests are more prominent.  Moreover, it assumes knowing the group membership of each sample at unlearning time, which is highly impractical in realistic scenarios.
In contrast, we introduce the first approach that handles unbalanced forget requests \textit{without} group membership annotations, while also being the first to address this problem for MLLMs. %

\input{tables/benchmark}

\paragraph{Benchmarking Unlearning.}
Existing facial datasets such as UTKFace~\cite{zhifei2017cvpr} and FairFace~\cite{karkkainen2021fairface} are designed for classification of demographic attributes, limiting profiles to a single image and a narrow set of attributes.  
However, as these established datasets are public, there is no control for their leakage in MLLM training data, limiting oversight of the source of information and model exposure\cite{mainitofu, mabenchmarking}.
To this end, TOFU~\cite{mainitofu} is the first benchmark that assesses unlearning effectiveness on LLMs under the right to be forgotten, by first fine-tuning them on fictitious identities, and then unlearning a few identities.
CLEAR~\cite{dontsov-etal-2025-clear} extends it to the multimodal domain by proposing to generate pictures of faces corresponding to fictitious identities in TOFU.
Similarly, MLLMU~\cite{liu2025protecting} and FIUBench~\cite{mabenchmarking} exploit image synthesis techniques and sampling of personal information to build realistic multimodal unlearning requests. 
Compared to previous efforts: (i) \ourbenchmark{} goes beyond facial datasets by providing multiple images per identity with multimodal Q\&A, and (ii) focuses on complete controllability of fictitious identity generation, allowing for a balanced VQA dataset for model fine-tuning and for controlled, unbalanced, and more realistic unlearning requests.%

%% file: tables/benchmark.tex
\begin{table}[tp]
    \caption{\textbf{\ourbenchmark{} \vs other unlearning benchmarks.} \ourbenchmark{} is more than $10\times$ bigger than existing benchmarks, while allowing realistic \textit{unbalanced unlearning}. 
    }
    \label{tab:benchmark}
    \scriptsize
    \centering
    \begin{tabularx}{\linewidth}{
        L{2.5cm}
        *{2}{Y}
        *{3}{C{1.5cm}}
    }
        benchmark & ids & imgs & Q\&A &
        \makecell[c]{unbalanced\\requests} &
         \makecell[c]{visual\\attrs.} \\
    \toprule
        MMUBench~\cite{li2024single} & 20 & 1k & 50 & \crossm & \crossm \\
        CLEAR~\cite{dontsov-etal-2025-clear} & 200 & 8k & 4k  & \crossm & \crossm \\
        MLLMU-Bench~\cite{liu2025protecting} & 500 &  1k & 10k  & \crossm & \crossm \\
        FIUBench~\cite{mabenchmarking} & 400 &  400 & 8k & \crossm & \crossm \\
        \textbf{\ourbenchmark{}} & \textbf{4k} & \textbf{40k} & \textbf{225k} & \checkm & \checkm \\
    \end{tabularx}
\end{table}

%% file: sec/3_problem.tex
\section{Problem formulation}
\label{subsec:problem}
Let $\model$ be a multimodal LLM parametrized by $\weights$, %
and trained on a dataset $\dataset[tr] = \{(\sample[i], \target[i])\}_{i=1}^{\population[tr]}$, where $\sample[i]$ is a user query (\ie, image and text) and $\target[i]$ its ground truth answer in the VQA setting.
The goal of machine unlearning is to remove the influence of a given \textit{forget} set $\dataset[f]\subset\dataset[tr]$ from the trained model, while preserving the information related to the retain set $\dataset[r] = \dataset[tr] \setminus \dataset[f]$.  

We consider the setting in which the forget samples, $(\sample[f],\target[f]) \in\dataset[f]$, contain personal information in the target answers $\target[f]$ (\eg, income, political orientation).
Given model weights and data, an effective unlearning algorithm outputs weights $\weights[u]$, such that the updated $\model[u]$ cannot correctly infer the personal information in $\dataset[f]$ better than chance level, \ie, $\forall (\sample[f],\target[f]), \model[u](\sample[f]) \neq \rephrasingf[{\target[f]}]$ up to rephrasing $\rephrasingf[{\cdot}]$,  while maintaining the original knowledge on $\dataset[r]$.

In realistic RTBF requests, the forget set consists of non-uniformly distributed users' data across demographic groups~\cite{epstein2020markers,cvrcek2006study},  requesting to be forgotten.
We define groups as the specific combination of \targetattr{target} and \protectedattr{protected} attributes. 
In particular, let $\targetspace$ be the set of \textit{target} attributes, \ie, those explicitly learned as the target $\target$ during fine-tuning, (\eg, the net-worth). Conversely, we refer to \textit{protected} attributes $\attributespace$ as those unavailable at the target level (\ie, not explicitly included in $\target$ during training), but that can still be inferred from the visual modality (e.g., ethnicity, gender). Hence, groups are formally defined as $\groupspace: \targetspace\times\attributespace$. %
As unlearning unbalanced forget sets can lead to biased models~\cite{de2025group}, unlearning algorithms should also mitigate biases against the dominant group in the forget set, in addition to addressing classical machine unlearning challenges.
Under Demographic Parity (DP) \cite{kusner2017counterfactual}, the unlearned model is expected to predict the target attribute independently of the protected attribute: $P(\hat{Y}=\target\mid A=\attribute) = P(\hat{Y}=\target\mid A=\neg\attribute)$,
where $\hat{Y}$ and $A$ are random variables describing the unlearned model's prediction and the protected attribute. 
In practice, the goal is two-fold: while the model should act as if the unlearned data never existed (\ie, not recalling unlearned identities), unlearning should not lead predictions to be \textit{biased} by protected attributes due to unbalanced forget requests.

%% file: sec/4_benchmark.tex
\begin{figure*}[tp]
    \centering
    \includegraphics[width=1.\linewidth]{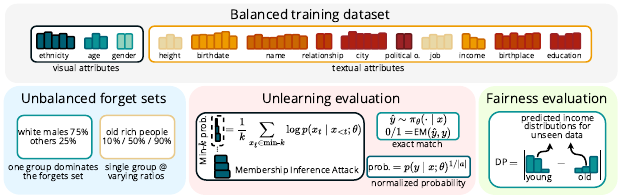}
    \caption{\textbf{\ourbenchmark{} evaluation protocol.} Identities contained in \ourbenchmark{} consist of visually and textually learned attributes. Attributes are balanced with minor variations to account for realistic attribute distributions. Additionally, \ourbenchmark{} benchmarks unlearning via unbalanced unlearning requests over two scenarios, \ie, with a group that partly dominates and with a group that totally dominates the forget set. Unlearning effectiveness is estimated via three metrics: membership inference attack, exact attribute match, and normalized answer probability. Finally, \ourbenchmark{} evaluates MLLM's fairness by computing the demographic parity on unseen data.
    }
    \label{fig:dataset}
\end{figure*}

\section{\ourbenchmark{}} 
This section details \ourbenchmark{}, the first multimodal benchmark for unbalanced unlearning.
\ourbenchmark{} is about 10$\times$ bigger than existing VQA unlearning benchmarks (see \cref{tab:benchmark}), featuring 4k identities and over 225k Q\&A pairs spread over 13 profile attributes covering textual and visual modalities (\cref{fig:teaser} right).

\noindent\paragraph{Visual modality.}
\label{subsec:visual_mod}
Images are obtained starting from 450k unconditioned StyleGAN2~\cite{karras2019style} generated faces.
As we want equal representation of gender, age, and ethnicity in the benchmark, we annotate each image with the FairFace~\cite{karkkainen2021fairface} model.
This allows us to subsample 4k identity images while preserving balance among attributes.
To simulate intra-identity variability, each image is augmented 10 times using Arc2Face~\cite{papantoniou2024arc2face}.
As original annotations can change after augmenting images, we apply a second round of annotation with FairFace, and choose attributes based on majority voting.
This results in 40k pictures with balanced attributes.
Identities and corresponding images are finally evenly split into two subsets, one used for training and unlearning (\ie, unlearning split), and the other for probing MLLMs' fairness (\ie, fairness split).

\noindent\paragraph{Textual modality.}
\label{subsec:text_mod}
Starting from unlearning split identities, we construct rich profiles consisting of 10 additional personal attributes (\eg, name, occupation, residence). 
Rather than directly relying on generative models~\cite{liu2025protecting}, attributes are assigned through a rule-based and stratified sampling procedure, ensuring balanced distributions and bias mitigation (see the Appendix). 
While certain attributes, such as names or locations, may be correlated with protected attributes, others (\eg, profession) are independent. 
From these profiles, we generate a large pool of Q\&A pairs. Contrary to previous approaches that sample questions via LLM prompting~\cite{liu2025protecting, mabenchmarking}, we generate attribute-specific templates with DeepSeek~\cite{guo2025deepseek} (\eg, ``In what city does \smalltt{<name>} live in?'', ``Where is this individual from?'') and populate them with identity-specific values. 
This allows us to synthesize novel Q\&As at scale, covering information about identities while maintaining control over diversity, correctness, and fairness. 
Thus, for each identity and attribute, we create 10 unique Q\&A instances for training and 2 for testing, which are obtained by rephrasing the training ones.
This results in approximately 225k pairs.
Finally, we restrict training Q\&As to textual attributes that cannot be inferred from images, while relying on the visual domain for directly observable characteristics, \ie, age, gender, and ethnicity (\Cref{fig:dataset}, top).
Instead, the test set features both Q\&As about visual and textual domains.

\subsection{Evaluation pipeline}
\label{subsec:eval_pipeline}
\Cref{fig:dataset} shows an overview of \ourbenchmark{} evaluation pipeline. 
\ourbenchmark{} identities include visually and textually learned attributes, balanced to reflect realistic distributions (grey box). 
Thus, MLLMs undergo fine-tuning on the \textit{unlearning split}, so unlearning can be performed on a subset of the learned identities.
Out of the 10 identity images in the \textit{unlearning partition}, five are used during fine-tuning and unlearning, and the remaining are reserved for evaluation \textit{only}, testing model robustness in answering rephrased questions on held-out profile images.

Unlearning is performed under two scenarios: single- and multi-group unlearning. 
In the former, a single group (\eg, old rich people in the image) dominates the forget set, whereas in the latter, one group accounts for 75\% (\eg, white males in the image) of the unlearning requests, with the remaining 25\% uniformly sampled across all groups. 
For the single-group setting, we unlearn varying ratios of the dominant group (\ie, 10\%, 50\%, 90\%).

Forgetting quality is established via three metrics: %
(i) \textit{\textbf{Membership Inference Attack}} (MIA)~\cite{shokri2017membership,carlini2022membership} evaluates whether the forget set has been removed by checking if the model’s output distributions on forget samples are distinguishable from those on members' data, with higher discriminability indicating better unlearning.
We compute the MIA score using the Min-K++~\cite{zhang2025min} method, which aggregates the log-likelihoods of the top-k tokens with lowest-probability to assess membership.
(ii) \textit{\textbf{exact match}} (EM) {evaluates the accuracy performance of the model on queried attributes, by accounting for the presence of the ground truth in the predicted answer (higher values indicate better retention, while lower values indicate better forgetting), and}
(iii) \textit{\textbf{normalized probability}} (prob.), which evaluates the length-normalized likelihood assigned to the correct answer, thereby quantifying the model’s confidence~\cite{mainitofu} on the queried information.

For assessing fairness, we evaluate the model on the unseen identities of the \textit{fairness partition} using \textit{\textbf{demographic parity}} (DP) \cite{kusner2017counterfactual}. 
Leveraging unseen data is crucial to decouple fairness assessment from memorization effects, ensuring that any measured bias reflects the model’s behavior rather than residual knowledge of the unlearning set (see the Appendix for details on the evaluation metrics). Finally, to ease the interpretation of the metrics with competing objectives, we follow previous works \cite{jia2023model, fan2023salun} and compute the \textit{\textbf{average gap}} (\avggap{}) as the average of per-metric difference (scaled by 100) with the gold standard, \ie, a model with unlearning performance equal to the retrained model without forget data and not accounting for protected attributes for its prediction ($\text{DP}=0$).

%% file: sec/5_method.tex
\section{\ourmethodlong{} (\ourmethod{})}
\label{sec:method}
To address all requirements of the unbalanced unlearning setting (\cref{subsec:problem}), we introduce \ourmethodlong{} (\ourmethod{}), an activation steering strategy that explicitly accounts for unbalanced forget sets.
\Cref{sec:method:steering} and \cref{sec:method:unlearning} introduce activation steering and how it can be used for permanent machine unlearning.
Instead, \cref{sec:method:bias} shows how to attenuate group information during unlearning, effectively mitigating {model} biases.

\begin{figure*}[tp]
    \centering
    \includegraphics[width=\linewidth]{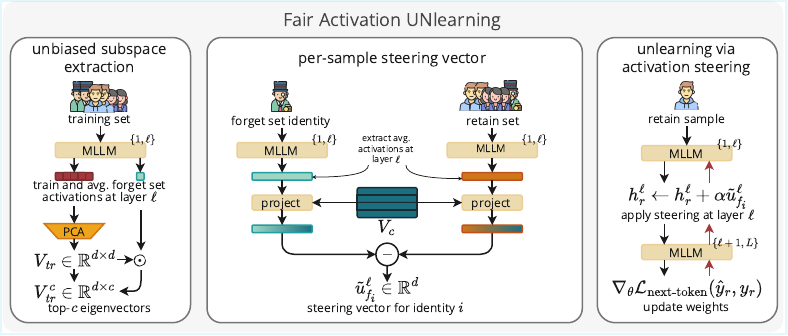}
    \caption{\textbf{\ourmethod{} overview.} {Left: Starting from the training set, \ourmethod{} extracts training samples activations and the \textit{average} forget set activation at layer $\ell$. Then, it performs the PCA on training activations, selecting components via dot product with the average forget set activations (see \cref{sec:method:bias}).} Center: We compute the unbiased steering vector for each sample in the forget set as the difference between the projected forget identity and retain vectors (see \cref{sec:method:unlearning}). Right: Permanent unlearning is performed via gradient descent on the retain set steered with forget identity vectors (see \cref{sec:method:unlearning}).}
    \label{fig:method}
\end{figure*}

\subsection{Preliminaries: activation steering}
\label{sec:method:steering}
Activation steering~\cite{rimsky2024steering,chen2025persona} changes the default behavior
of (multimodal) large language models $\model$ towards a desired one.
Under linear assumptions of the MLLM space~\cite{parklinear,turner2023steering}, steering alters the model output to align with the desired behavior by adding a displacement vector to intermediate activations.
Formally, consider $\bdata[+]=\{\bsample[+]^{(j)}\}_{j=1}^{\bcardinality[+]}$ and 
$\bdata[-]=\{\bsample[-]^{(j)}\}_{j=1}^{\bcardinality[-]}$ two datasets associated with positive and negative behaviors respectively (\eg, safe answers \vs unsafe ones), and let $\hidden[b] \in\mathbb{R}^{d}$ be the residual stream activation at layer $\ell$, $\ell \in \{\range{1}{L}\}$, for the sample $\bsample[]$. 
We extract $\avghidden[+]$, \ie, the average activation at layer $\layer$, for a positive behavior we want to elicit from the model, and  $\avghidden[-]$ as its negative counterpart:
\begin{equation}
\avghidden[+] = \frac{1}{\bcardinality[+]}\sum_{b\in \bdata[+]} \hidden[b],\quad \avghidden[-]=\frac{1}{\bcardinality[-]}\sum_{b \in \bdata[-]} \hidden[b].  
\end{equation}
Then, the activation steering vector $\steering$ is computed as the difference between the positive and negative vectors, while at inference time, the positive behavior for a sample $b$ 
can be elicited by subtracting the scaled steering vector:
\begin{subequations}
\begin{center}
\begin{minipage}{0.4\textwidth}
\begin{equation}
\label{eq:steering}
    \steering = \avghidden[-] - \avghidden[+],
\end{equation}
\end{minipage}
\begin{minipage}{0.4\textwidth}
\begin{equation}
\label{eq:act_steering}
    \hidden[b] \leftarrow \hidden[b] - \coeff\steering.
\end{equation}
\end{minipage}
\end{center}
\end{subequations}
After steering the activation at layer $\ell$, the forward pass continues until the last layer. 
Since tokens generated by layers $m>\ell$ are influenced by the steering, the final output of the model changes as well, altering its behavior.

\subsection{Permanent unlearning via activation steering}
\label{sec:method:unlearning} 
We apply activation steering to machine unlearning by treating retain-set activations as positive behavior and forget-set activations as negative.
Thus, \cref{eq:act_steering} becomes $\hidden \leftarrow \hidden - \coeff\steering[f]$, where $\steering[f] = \avghidden[f]-\avghidden[r]$;
$\avghidden[r]$ and $\avghidden[f]$ are the retain and forget \textit{average} activations, and $\hidden$ is the sample activation.
Intuitively, steering makes forget queries behave like retain ones, yielding incorrect responses.

However, when used for unlearning, steering has three main problems. 
First, it induces only a temporary change in activations \textit{without} removing the underlying knowledge, which contrasts with the RTBF and the core motivation behind unlearning.
Second, steering should be selectively applied to forget samples only, as steering retain data would distort the preserved knowledge, degrading its performance. %
Finally, although inference-time steering can achieve some degree of unlearning, it does not scale well to a large number of identities, since as the forget set grows, the average activation $\avghidden[f]$ approaches the global mean.

\paragraph{Weight updates via activation steering.} 
To overcome activation steering limitations, we adopt a different approach: rather than steering the model's output at inference time, we make the MLLM robust against activations that could leak personal data (\ie, queries about identities to be forgotten) through gradient updates.
To this end, we iterate over the retain set and unlearn by emulating forget samples via steering retain activations toward those of the forget set. Hence, steered activations emulating forget samples are associated with wrong answers (the retain samples ground-truth) and the weights are optimized via gradient descent.
Let $(\sample[r], \target[r])\in\dataset[r]$ be a retain set sample, and $\dataset[{f_i}]$ the subset of the forget set belonging to the identity $i$. We can steer the retain sample as:
\begin{subequations}
\begin{center}
\begin{minipage}{0.4\textwidth}
\begin{equation}
\steering[f_i] = \avghidden[f_i]-\avghidden[r],
\label{eq:steering_ul}
\end{equation}
\end{minipage}
\begin{minipage}{0.4\textwidth}
\begin{equation}
\hidden[r] \leftarrow \hidden[r] + \coeff\steering[f_i].
\label{eq:act_steering_ul}
\end{equation}
\end{minipage}
\end{center}
\end{subequations}
Compared to the previous formulation, steering is computed on the activation $\hidden[r]$ of a retain sample (rather than any sample) by adding the negative behavior induced by the average activation of the $i$-th identity in the forget set, $\avghidden[f_i]$.
Conceptually, \cref{eq:steering_ul,eq:act_steering_ul} displace retain activations toward those of a \textit{specific forget identity}, functioning as a pseudo-rehearsal mechanism~\cite{shin2017continual,atkinson2021pseudo}. {This mechanism does not directly use individual activations of the forget set but instead relies on their average representation, thereby enforcing robustness against forget set activations.}
Minimizing the next-token prediction loss $\mathcal{L}_\text{next-token}(\pred[r], \target[r])$ over the steered output $\pred[r]$ forces the model to be robust against forget set activations.
Thus, the model is permanently unlearned.

\subsection{Bias-informed PCA}
\label{sec:method:bias}
As the forget set is likely unbalanced toward certain demographic groups in realistic RTBF requests, directly applying \cref{eq:act_steering_ul} can lead the MLLM toward learning spurious correlations that bias the model (see \cref{subsec:ablations}).
To mitigate this effect, unlearning should only concern the main directions associated with the forget data \textit{without} altering those related to group knowledge.
Applied to the steering strategy, we propose to remove group information from $\avghidden[f_i]$ and $\avghidden[r]$ while preserving components that discriminate identities.
To identify such directions, we  %
perform Principal Component Analysis (PCA) on the model's activations and remove components that are mostly aligned with the group information.

Formally, let matrix $\allhidden[tr]\in\mathbb{R}^{\population[tr]\times d}$ represent all training set activations at layer $\ell$, centered by subtracting the average training set activation $\avghidden[tr]$.
We thus compute their covariance matrix $\covariance[tr]$, to calculate the PCA: $\covariance[tr] = \components\eigenvalues\components[\top]$, where $\components\in\mathbb{R}^{d\times d}$ and $\eigenvalues\in\mathbb{R}^{d\times d}$ (diag) are orthonormal eigenvectors and their eigenvalues. 
Now consider forget set activations, since realistic RTBF requests are assumed to be unbalanced toward specific groups~\cite{bertram2019five,epstein2020markers}, we expect the them to exhibit \textit{limited variance} with respect to group-related information.
Consequently, the average activation of the forget set should serve as a reasonable estimate of the group information in the activation space.
Given this assumption, we center the average forget set activation $\avghidden[f]\in\mathbb{R}^d$ and project it to the eigenvector space:
\begin{equation}
    \label{eq:pca_get_info}
    \avghidden[fp] = \components[\top](\avghidden[f] - \avghidden[tr]).
\end{equation}
Intuitively, each scalar in the projected activation $\avghidden[fp]$ identifies the importance of each eigenvector \wrt the dominant unlearning group.
Therefore, keeping only the $\topcomponents$ components of $\components$ with the smallest magnitudes yields an unbiased subspace {$\components[\topcomponents]\in\mathbb{R}^{d\times c}$}, effectively informing the selection of components based on the bias.
The unbiased forget and retain activations are then computed as:
\begin{equation}
    \label{eq:project_unbias}
    \unbiasedhidden[f_i] = \components[\topcomponents]\components[\topcomponents\top] (\avghidden[f_i]-\avghidden[tr]) + \avghidden[tr],\quad
    \unbiasedhidden[r] = \components[\topcomponents]\components[\topcomponents\top](\avghidden[r]-\avghidden[tr]) + \avghidden[tr],
\end{equation}
{where the average forget identity activation $\unbiasedhidden[f_i]$ and retain activation $\unbiasedhidden[r]$ can be used to compute the unbiased steering vector and, consequently, the unbiased activation steering at unlearning time:}

\begin{subequations}
\begin{center}
\begin{minipage}{0.4\textwidth}
\begin{equation}
\label{eq:steering_unl_unbiased}
    \unbiasedsteering[f_i] = \unbiasedhidden[f_i] - \unbiasedhidden[r],
\end{equation}
\end{minipage}
\begin{minipage}{0.4\textwidth}
\begin{equation}
\label{eq:act_steering_unl_unbiased}
    \hidden[r] \leftarrow \hidden[r] + \coeff\unbiasedsteering[f_i].
\end{equation}
\end{minipage}
\end{center}
\end{subequations}
Notably, when forget requests are balanced across groups, the forget-set average does not encode any dominant group signal. 
As a consequence, the projected activations from \cref{eq:project_unbias} primarily capture identity-specific structure
resulting in stable model performance even under balanced regimes (see \cref{tab:fiubench}).

\paragraph{Putting all together.}
\Cref{fig:method} provides an overview of \ourmethod{}.
Before unlearning, we collect activations from the training set and perform the PCA.
Components are ranked based on the average activation of the forget set projected onto the eigenvector space (see \cref{eq:pca_get_info} and \cref{fig:method} left), with larger magnitudes indicating greater relevance of the corresponding components to the group.
Next, we compute the average activation of the retain set at layer $\ell$, as well as the mean activation for each identity in the forget set.
Projecting these vectors onto the unbiased eigenvector subspace and mapping them back to the original space (see \cref{eq:project_unbias}) produces the unbiased steering vector for each ID (see \cref{eq:steering_unl_unbiased} and \cref{fig:method} center).
During unlearning, we iterate over the retain set and, for each batch, we sample a forget steering vector to modify the intermediate activation of the retain samples (see \cref{eq:act_steering_unl_unbiased} and \cref{fig:method} right).
Repeating this process over multiple iterations ensures that the model becomes robust to forget set activations, thereby preventing leakage of private information (see Appendix).

%% file: sec/6_experiments.tex
\section{Experiments}
\input{tables/qwen_main}
\label{sec:experiments}
We evaluate \ourmethod{} on \ourbenchmark{}, which stress-tests models under unbalanced unlearning, and on the established FIUBench~\cite{mabenchmarking}, providing a reference with prior literature. Additionally, we assess models' general utility in the VQA task via the widely used MME~\cite{fumme} and ablate on the key design choices. %

\paragraph{Unlearning scenarios.}
We evaluate multiple \protectedattr{protected}-\targetattr{target} attribute pairs (different groups): \protectedattr{man} with \targetattr{right-wing} political orientation, \protectedattr{adult} with \targetattr{low} net worth, \protectedattr{asian}–\targetattr{single}, and \protectedattr{woman}–\targetattr{tall}. We report results for the first two pairs in the main text and defer the rest to the Appendix. To analyze unlearning effects on fairness, we explore the two complementary setups defined in \cref{subsec:eval_pipeline}, \ie, with a single group dominating the forget set, and with a group dominating 75\% of the forget request while uniformly sampling the remaining groups.

\paragraph{Baselines.} We compare \ourmethod{} with a comprehensive set of baselines, including gradient ascent (GA)~\cite{thudi2022unrolling}, gradient difference (GAD)~\cite{liu2022continual}, random labeling (RL)~\cite{Golatkar_2020_CVPR}, the rejection-based ``I don't know'' fine-tuning (IDK)~\cite{mainitofu}, the unlearning state-of-the-art SimNPO~\cite{fan2024simplicity}, FTTP~\cite{li2025forget} and the activation steering strategy LUNAR~\cite{shen2025lunar}. 
We also report results for the original model and for a model retrained from scratch without the forget data. 
We adapt MIU to MLLMs by using the average embedding from the last block and consider it a privileged baseline because it relies on auxiliary group annotations.
For a fair comparison, all methods are implemented on \qwen~\cite{bai2025qwen2} and \idefics~\cite{laurenccon2024building}, and the computational budget is fixed for all experiments.

\subsection{Results in unbalanced unlearning}
\label{sec:exp:unbalanced}

\paragraph{Main comparison.} \Cref{tab:qwen_main,tab:idefics_main} reports on unbalanced unlearning in \ourbenchmark{}.
In both tables, we unlearn 50\% of the groups defined above, respectively for \qwen and \idefics. 
When using \qwen as the MLLM (\cref{tab:qwen_main}), \ourmethod{} achieves the best trade-off between unlearning effectiveness, model utility, and fairness as the \avggap{} shows in both settings. The comparison with MIU highlights the challenging balance between these conflicting objectives: despite relying on group labels, MIU generally improves forgetting at the cost of worse DP. We hypothesize MIU struggles to generalize group annotations to unseen identities in the DP evaluation. With no group annotations, excluding IDK that overforgets, resulting in insufficient retrain exact match (EM), \ourmethod{} attains the lowest forget EM across the two settings (34.83\% \vs 39.77\% of RL), the best normalized probability (0.411 \vs 0.419 of SimNPO), and the lowest demographic parity (0.18 \vs 2.51 of RL).
However, it slightly lags in model utility (57.59\% \vs 61.77\% of SimNPO), \ie, retain EM, and normalized probability (0.490 \vs 0.507 of IDK). 
Nonetheless, methods that achieve DP comparable to \ourmethod{} generally perform worse on the retain and forget set metrics (\eg, RL when unlearning \protectedattr{man} with \targetattr{right-wing} political orientation). Compared with LUNAR, our approach yields more stable results, whereas the direct competitor either fails to unlearn effectively or refuses to answer all identity-related questions.
Finally, in terms of MIA, we note that \ourmethod{} achieves competitive results with the state-of-the-art SimNPO (55.10 \vs 55.50 of SimNPO) and RL (55.10 \vs 53.70). 
Similar trends are also observed with \idefics as underlying MLLM (\cref{tab:idefics_main}).
\ourmethod{} strikes a good balance between unlearning and fairness, achieving similar unlearning performance as SimNPO, while yielding a fairer unlearning process (6.35 \vs 25.35 DP).   
\input{tables/idefics_main}

\paragraph{Results at varying unlearning ratios.}                         
\begin{figure}[t]
    \centering

    \begin{tikzpicture}
        \node[anchor=south west,inner sep=0] (image) at (0,0)
            {\includegraphics[width=0.85\linewidth]{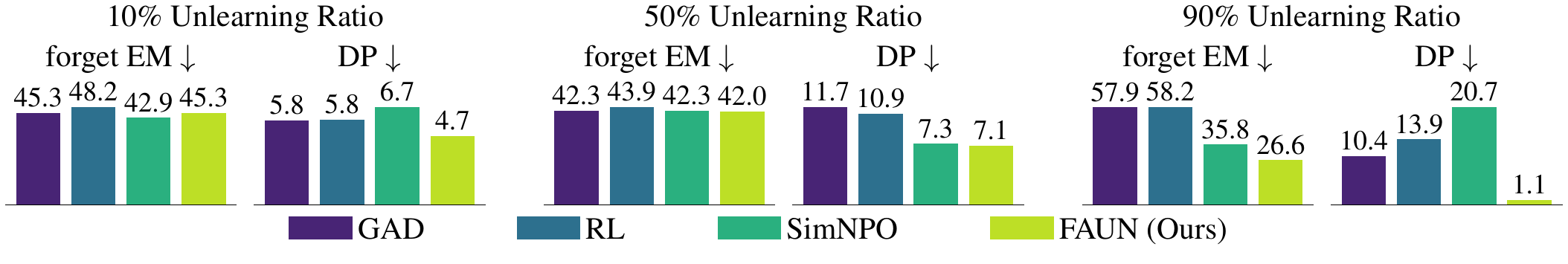}};
            
        \begin{scope}[x={(image.south east)},y={(image.north west)}]
            \node[font=\fontsize{6}{7}\selectfont] at (0.288,0.132) {\cite{liu2022continual}};
            \node[font=\fontsize{6}{7}\selectfont] at (0.416,0.132) {\cite{Golatkar_2020_CVPR}};
            \node[font=\fontsize{6}{7}\selectfont] at (0.590,0.132) {\cite{fan2024simplicity}};
        \end{scope}
    \end{tikzpicture}
    
    \caption{\textbf{Unbalanced unlearning at different ratios on \qwen.} Performance of different methods in terms of forget exact match and demographic parity on \qwen where the forget set is composed of 10\%, 50\% and 90\% of all datapoints from the \protectedattr{adult} with \targetattr{low} net worth single-group.}
    \label{fig:unlearning-ratio}
\end{figure}
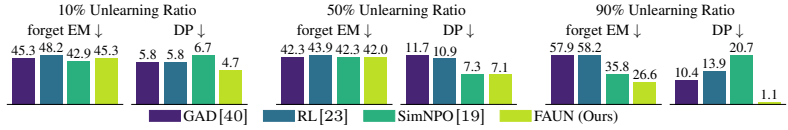
\cref{fig:unlearning-ratio} reports forget EM and DP for \qwen when unlearning 10\%, 50\% and 90\% of \protectedattr{adult} with \targetattr{low} net-worth.
\ourmethod{} achieves the best forget EM in two out of the three considered ratios, with a larger advantage in the challenging 90\% unlearning ratio experiment (26.59\% \vs 35.82\% of SimNPO). 
DP confirms that disentangling group information through our bias-informed PCA consistently yields fairer models after unlearning compared to competing methods.
Notably, \ourmethod{} sharply reduces the demographic parity to 1.07 while scoring the lowest forget EM (\ie, 26.59) at 90\% unlearning ratio. 
Overall, \ourmethod{} achieves the best balance between unlearning effectiveness and fairness, consistently preserving model utility across varying unlearning ratios.
We refer to the Appendix for additional results.

\paragraph{Results in the multi-group setting.}
\cref{fig:mixed} examines the multi-group setting, in which unlearning targets 10\% of all identities, 75\% of which belong to the dominant group, \ie, \protectedattr{man} with \targetattr{right-wing} political orientation. 
We report both retain and forget EM and DP. 
Our method attains the second-best forgetting quality (34.83\%), slightly behind GAD (29.96\%). 
Yet, GAD exhibits substantial degradation in knowledge retention (55.59\% \vs 59.51\% of \ourmethod{}). 
On retain data, \ourmethod{} outperforms approaches that achieve similar forgetting strength (\eg, 59.51\% \vs 58.08\% of SimNPO), while notably delivering large fairness gain over SimNPO (8.15 \vs 20.22), despite exhibiting similar unlearning efficacy. 
While methods that over-forget (GAD) or do not forget at all (RL) may show marginal advantages on isolated axes, \ourmethod{} compares favorably overall, further supporting the need to account for the forget set imbalance while unlearning.
\input{tables/fiubench}
\input{tables/utility}

\paragraph{Results on FIUBench.}
\Cref{tab:fiubench} compares \ourmethod{} with existing baselines on FIUBench~\cite{mabenchmarking}, an established multimodal unlearning benchmark. This experiment demonstrates our method competitiveness even when the forget set is uniformly distributed.
\ourmethod{} shows great performance, achieving the lowest forget EM (25.09\% \vs 27.87\% of GAD), while scoring a retain EM comparable with other methods.
Particularly, \ourmethod{} falls slightly behind compared to RL in retain EM (44.53\% \vs 46.84\%); yet, it gets a far better forget EM (25.09\% \vs 31.90\%), highlighting its superior unlearning performance.
Finally, \ourmethod{} also shows the best MIA score across all algorithms (54.10 \vs 53.90) and lowest avg.~gap.

\paragraph{Results on VQA utility.} 
To probe general utility on VQA, we compare accuracy on the MME~benchmark \cite{fumme} after unlearning 50\% of right-wing men. Both the \textbf{C}ognition and \textbf{P}erception splits are considered.
Since the retain set only contains \ourbenchmark{} data, all unlearning methods degrade utility.
Notably, \ourmethod{} achieves competitive utility while showing the best forgetting performance (\cref{tab:qwen_main}).
\ourmethod{} yields on-par results with SimNPO, with a minor advantage in Cognition (478.6 \vs 485.7), and a disadvantage in Perception (1276.0 \vs 1249.0).

\input{tables/ablations}
\subsection{Ablations}
\label{subsec:ablations}
\Cref{tab:ablations} shows \ourmethod{} component ablations on \qwen when unlearning 50\% of \protectedattr{man} with \targetattr{right-wing} political orientation.
From left to right, we list permanent unlearning via activation steering (\cref{eq:act_steering_ul}), a na\"ive PCA for debiasing (details in Appendix), and \ourmethod{}, where component selection is performed via \cref{eq:act_steering_unl_unbiased}.
Activation steering alone yields the lowest forget EM (31.44\%) but the highest DP (23.32). 
Using a na\"ive PCA for debiasing slightly improves DP (18.28) but worsens forgetting (38.66\%), indicating poor disentanglement between identity and group information. 
In contrast, a bias-informed PCA effectively removes group bias (0.16 in DP), enabling fair unlearning (34.83\%) with strong utility (57.59\% retain EM). 
In the Appendix we ablate on the number of components $c$.

%% file: tables/qwen_main.tex
\begin{table}[tp]
    \centering
    \caption{\textbf{Unbalanced unlearning on \qwen.} Performance of different methods where the forget set is composed of 50\% of all datapoints of a single group.}
    \resizebox{0.8\linewidth}{!}{%
    \scriptsize
    \begin{tabularx}{\linewidth}{
        L{1.5cm}
        *{7}Y
    }
    \multirow{2.5}{*}{method} & \multicolumn{2}{c}{forget} & \multicolumn{2}{c}{retain} & \multirow{2.5}{*}{MIA$\uparrow$} & \multirow{2.5}{*}{DP$\downarrow$} & \multirow{2.5}{*}{avg.~gap$\downarrow$}\\
\cmidrule(lr){2-3}
\cmidrule(lr){4-5}
    & EM~$\downarrow$ & prob.~$\downarrow$ & EM~$\uparrow$ & prob.~$\uparrow$ \\
\toprule
    \multicolumn{7}{l}{\textcolor{gray}{\textit{\protectedattr{man} with \targetattr{right-wing} political orientation}}} & \\
    original  & 74.91 &	0.480 &	67.24 &	0.492 &	52.05 &	4.25 &	 \cellcolor{experiment}{-}\\
    retrained & 17.20 &	0.270 &	67.63 &	0.490 &	65.09 &	30.42 & \cellcolor{experiment}{-} \\
\cmidrule(lr){1-8}
\textcolor{gray}{MIU~\cite{de2025group}} &	\textcolor{gray}{16.93} &	\textcolor{gray}{0.234} &	\textcolor{gray}{68.09} &	\textcolor{gray}{0.419} &	\textcolor{gray}{41.29} &	\textcolor{gray}{13.62} &	\cellcolor{experiment}{\textcolor{gray}{8.14}} \\
\hdashline
    GA~\cite{thudi2022unrolling} & 63.13 &	0.431 &	57.53 &	0.448 &	51.99 &	1.24 &	\cellcolor{experiment}{15.12} \\
    GAD~\cite{liu2022continual} & 40.44 &	0.428 &	58.04 &	0.482 &	53.83 &	16.35 &	\cellcolor{experiment}{12.85} \\
    IDK~\cite{mainitofu} & 61.53 &	0.459 &	74.77 &	0.507 &	53.94 &	13.76 &	\cellcolor{experiment}{16.17} \\
    RL~\cite{Golatkar_2020_CVPR} & 39.77 &	0.422 &	58.96 &	0.479 &	53.65 &	2.51 &	\cellcolor{experiment}{\underline{10.26}} \\
    FTTP~\cite{li2025forget} & 40.14 &	0.413 &	54.46 &	0.463 &	53.53 &	5.26 &	\cellcolor{experiment}{11.67} \\
    SimNPO~\cite{fan2024simplicity} & 42.79 &	0.419 &	61.77 &	0.486 &	55.49 &	6.61 &	\cellcolor{experiment}{10.50} \\
    LUNAR~\cite{shen2025lunar} & 19.79 &	0.358 &	16.67 &	0.364 &	48.95 &	5.93 &	\cellcolor{experiment}{16.17} \\
    \textbf{\ourmethod{}}~\textbf{(Ours)} & 34.83 &	0.411 &	57.59 &	0.487 &	55.13 &	0.18 &	\cellcolor{experiment}{\textbf{8.72}} \\
    \\[-0.5em]
    \multicolumn{7}{l}{\textcolor{gray}{\textit{\protectedattr{adult} with \targetattr{low} net-worth}}} \\
    original & 74.57 &	0.483 &	67.23 &	0.492 &	49.64 &	4.35 & \cellcolor{experiment}{-} \\
    retrained & 17.05 &	0.287 &	66.26 &	0.491 &	63.39 &	17.65 & \cellcolor{experiment}{-} \\
\cmidrule(lr){1-8}
\textcolor{gray}{MIU~\cite{de2025group}} &	\textcolor{gray}{22.19} &	\textcolor{gray}{0.338} &	\textcolor{gray}{68.62} &	\textcolor{gray}{0.422} &	\textcolor{gray}{30.42} &	\textcolor{gray}{11.70}  &	\cellcolor{experiment}{\textcolor{gray}{10.70}} \\
\hdashline
    GA~\cite{thudi2022unrolling} & 51.39 &	0.412 &	46.81 &	0.421 &	49.34 &	7.16 &	\cellcolor{experiment}{15.74} \\
    GAD~\cite{liu2022continual} & 42.31 &	0.426 &	59.59 &	0.481 &	52.07 &	11.70 &	\cellcolor{experiment}{11.63} \\
    IDK~\cite{mainitofu} & 15.09 &	0.360 &	16.63 &	0.379 &	50.36 &	0.04 &	\cellcolor{experiment}{13.87} \\
    RL~\cite{Golatkar_2020_CVPR}& 43.93 &	0.437 &	64.12 &	0.492 &	52.74 &	10.91 &	\cellcolor{experiment}{10.96} \\
    FTTP~\cite{li2025forget} & 44.22 &	0.440 &	56.86 &	0.484 &	51.77 &	13.54 &	\cellcolor{experiment}{12.95} \\
    SimNPO~\cite{fan2024simplicity} & 42.31 &	0.429 &	60.51 &	0.491 &	54.21 &	7.30 &	\cellcolor{experiment}{\underline{10.28}} \\
    LUNAR~\cite{shen2025lunar} & 29.04 & 0.286 & 28.61 & 0.283 & 48.56 & 7.56 &	\cellcolor{experiment}{16.97} \\
    \textbf{\ourmethod{}}~\textbf{(Ours)}  & 41.97 &	0.425 &	63.84 &	0.490 &	53.42 &	7.10 &	\cellcolor{experiment}{\textbf{9.71}} \\
    \end{tabularx}
    }
    \label{tab:qwen_main}
\end{table}

%% file: tables/idefics_main.tex
\begin{table}[t]
    
    \centering
    \caption{\textbf{Unbalanced unlearning on \idefics.} Performance of different methods where the forget set is composed of 50\% of all datapoints of a single group.}
    \resizebox{0.8\linewidth}{!}{%
    \scriptsize
    \begin{tabularx}{\linewidth}{
        L{1.5cm}
        *{7}Y
    }
    \multirow{2.5}{*}{method} & \multicolumn{2}{c}{forget} & \multicolumn{2}{c}{retain} & \multirow{2.5}{*}{MIA $\uparrow$}& \multirow{2.5}{*}{DP $\downarrow$} & \multirow{2.5}{*}{\avggap{} $\downarrow$}\\
\cmidrule(lr){2-3}
\cmidrule(lr){4-5}
     & EM $\downarrow$ & prob.\ $\downarrow$ & EM $\uparrow$ & prob.\ $\uparrow$ &  &  \\
\toprule
    \multicolumn{7}{l}{\textcolor{gray}{\textit{\protectedattr{man} with \targetattr{right-wing} political orientation}}}\\
    original & 71.21 &	0.475 &	64.78 &	0.486 &	52.28 &	7.08 & \cellcolor{experiment}{-} \\
    retrained  & 17.94 &	0.272 &	60.77 &	0.465 &	63.98 &	22.00 & \cellcolor{experiment}{-} \\
\cmidrule(lr){1-8}
\textcolor{gray}{MIU~\cite{de2025group}} & \textcolor{gray}{29.50} & \textcolor{gray}{0.321} & \textcolor{gray}{63.03} & \textcolor{gray}{0.427} & \textcolor{gray}{32.57} & \textcolor{gray}{8.49} &	\cellcolor{experiment}{\textcolor{gray}{10.40}} \\
\hdashline
    GA~\cite{thudi2022unrolling} & 61.34 &	0.427 &	57.71 &	0.443 &	52.05 &	9.12 &	\cellcolor{experiment}{14.20} \\
    GAD~\cite{liu2022continual} & 30.39 &	0.363 &	51.63 &	0.431 &	54.63 &	2.55 &	\cellcolor{experiment}\underline{7.67} \\
    IDK~\cite{mainitofu} & 55.86 &	0.438 &	70.86 &	0.492 &	53.52 &	7.43 &	\cellcolor{experiment}{14.19} \\
    RL~\cite{Golatkar_2020_CVPR} & 20.84 &	0.366 &	37.11 &	0.422 &	54.45 &	15.61 &	\cellcolor{experiment}{10.91} \\
    FTTP~\cite{li2025forget} & 30.95 &	0.382 &	46.35 &	0.433 &	53.65 &	7.83 &	\cellcolor{experiment}{9.98} \\
    SimNPO~\cite{fan2024simplicity} & 33.97 &	0.399 &	53.60 &	0.467 &	55.93 &	25.19 &	\cellcolor{experiment}{11.55} \\
    LUNAR~\cite{shen2025lunar} & 69.79 &	0.473 &	65.05 &	0.487 &	52.23 &	8.53 &	\cellcolor{experiment}{16.44} \\
    \textbf{\ourmethod{}}~\textbf{(Ours)} & 29.78 &	0.359 &	50.83 &	0.438 &	56.08 &	1.02 &	\cellcolor{experiment}{\textbf{7.01}} \\
    \\[-0.5em]
    \multicolumn{7}{l}{\textcolor{gray}{\textit{\protectedattr{adult} with \targetattr{low} net-worth}}} \\
    original & 69.08 &	0.473 &	64.97 &	0.486 &	50.88 &	0.78 & \cellcolor{experiment}{-} \\
    retrained & 18.38 &	0.299 &	63.98 &	0.483 &	62.28 &	12.99 &\cellcolor{experiment}{-} \\
\cmidrule(lr){1-8}
\textcolor{gray}{MIU~\cite{de2025group}} & \textcolor{gray}{53.80} & \textcolor{gray}{0.457} & \textcolor{gray}{64.98} & \textcolor{gray}{0.480} & \textcolor{gray}{34.07} & \textcolor{gray}{12.80} &	\cellcolor{experiment}{\textcolor{gray}{15.56}} \\
\hdashline
    GA~\cite{thudi2022unrolling} & 62.14 &	0.460 &	61.75 &	0.478 &	51.08 &	1.12 &	\cellcolor{experiment}{12.49} \\
    GAD~\cite{liu2022continual} & 26.71 &	0.380 &	43.24 &	0.437 &	53.29 &	6.52 &	\cellcolor{experiment}{9.54} \\
    IDK~\cite{mainitofu}  & 57.17 &	0.443 &	71.33 &	0.491 &	53.15 &	15.81 &	\cellcolor{experiment}{14.38} \\
    RL~\cite{Golatkar_2020_CVPR} & 34.22 &	0.401 &	53.60 &	0.460 &	53.45 &	6.58 &	\cellcolor{experiment}{9.02} \\
    FTTP~\cite{li2025forget} & 37.05 &	0.409 &	53.42 &	0.457 &	52.27 &	10.04 &	\cellcolor{experiment}{10.48} \\
    SimNPO~\cite{fan2024simplicity} & 33.70 &	0.395 &	51.12 &	0.453 &	54.85 &	4.90 &	\cellcolor{experiment}\underline{8.86} \\
    LUNAR~\cite{shen2025lunar} & 68.27 &	0.474 &	65.12 &	0.487 &	50.81 &	0.02 &	\cellcolor{experiment}{13.40} \\
    \textbf{\ourmethod{}}~\textbf{(Ours)}  & 31.97 &	0.364 &	52.47 &	0.435 &	54.40 &	6.35 &	\cellcolor{experiment}{\textbf{8.44}} \\
    \end{tabularx}
    \label{tab:idefics_main}
    }
\end{table}

%% file: tables/fiubench.tex
\begin{table}[t]
    \centering
    \caption{\textbf{Unlearning on FIUBench.} Performance of different methods, results are computed on \qwen.}
    \resizebox{0.8\linewidth}{!}{%
    \scriptsize
    \begin{tabularx}{\linewidth}{
        L{1.5cm}
        *{6}Y
    }
    \multirow{2.5}{*}{method} & \multicolumn{2}{c}{forget} & \multicolumn{2}{c}{retain} & \multirow{2.5}{*}{MIA $\uparrow$} & \multirow{2.5}{*}{\avggap{} $\downarrow$}\\
\cmidrule(lr){2-3}
\cmidrule(lr){4-5}
    & EM $\downarrow$ & prob.\ $\downarrow$ & EM $\uparrow$ & prob.\ $\uparrow$ \\
\toprule
    original &  55.86 & 0.086 & 57.15 & 0.104 &    54.60 &    \cellcolor{experiment}{-} \\
    retrained & 7.57 &    0.049 &    55.64 &    0.113 &    57.20 &    \cellcolor{experiment}{-} \\
\cmidrule(lr){1-7}
    GAD~\cite{liu2022continual} & 27.87 &    0.108 &    41.84 &    0.131 &    50.20 &    \cellcolor{experiment}{8.13} \\
    RL~\cite{Golatkar_2020_CVPR} & 31.90 &    0.122 &    46.84 &    0.143 &    52.90 &    \cellcolor{experiment}{7.95} \\
    FTTP~\cite{li2025forget} & 39.22 & 0.130 & 46.25 & 0.110 & 54.00 &    \cellcolor{experiment}{8.77} \\
    SimNPO~\cite{fan2024simplicity} & 33.42 &    0.092 &    44.87 &    0.118 &    53.90 &    \cellcolor{experiment}{\underline{7.45}} \\
    \textbf{\ourmethod{}}~\textbf{(Ours)} & 25.09 &    0.108 &    44.53 &    0.137 &    54.10 &    \cellcolor{experiment}{\textbf{6.67}} \\
    \end{tabularx}
    \label{tab:fiubench}
    }
    
\end{table}

%% file: tables/utility.tex
\begin{table}[t!]
    \centering
    \caption{\textbf{VQA utility on \qwen{}.} Model utility on MME of different methods after unlearning 50\% of \protectedattr{man} with \targetattr{right-wing} political orientation.}
    \resizebox{0.8\linewidth}{!}{%
    \scriptsize
    \begin{tabularx}{\linewidth}{
        L{1.8cm}
        *{9}{Y}
    }
        benchmark & original & GAD & RL & SimNPO &\ourmethod{} \\
    \toprule
        MME-C $\uparrow$ & 671.8 & 246.4 & 375.3 & \underline{478.6}  &  \textbf{485.7} \\
        MME-P $\uparrow$ & 1462.0 & 854.7 & 1174.0 & \textbf{1276.0}  & \underline{1249.0} \\
    \end{tabularx}
    \label{tab:utility-mme}
    }
\end{table}

%% file: tables/ablations.tex
\begin{figure}[t]
    \centering
    
    \begin{minipage}[t]{0.48\textwidth}
        \caption{\textbf{\ourmethod{} ablations.} Ablations are computed on \qwen{}, by unlearning 50\% of \protectedattr{man} with \targetattr{right-wing} political orientation. From left to right, we report \cref{eq:act_steering_unl_unbiased}, \ie, permanent unlearning via activation steering, na\"ive PCA for debiasing, and \cref{eq:pca_get_info}, \ie, adding group information to select PCA components.}
        \label{tab:ablations}
        \resizebox{\linewidth}{!}{%
        \scriptsize
        \begin{tabular}{
        ccc ccc
    }
        \cref{eq:act_steering_ul} & PCA & \cref{eq:act_steering_unl_unbiased} & forget EM $\downarrow$ & retain EM $\uparrow$ & DP $\downarrow$ \\
        \toprule
        \checkm{} & \crossm{} & \crossm{} & 31.44 & 58.25 & 23.32 \\ 
        \checkm{} & \checkm{} & \crossm{} & 38.66 & 62.94 & 18.28 \\
        \checkm{} & \checkm{} & \checkm{} & 34.83 & 57.59 & 0.16 \\
    \end{tabular}
    }
    \end{minipage}
    \hfill
    \begin{minipage}[t]{0.48\textwidth}
        \centering
        
        \vspace{3pt}
        \begin{tikzpicture}
        \node[anchor=south west,inner sep=0] (image) at (0,0) {\includegraphics[width=\linewidth]{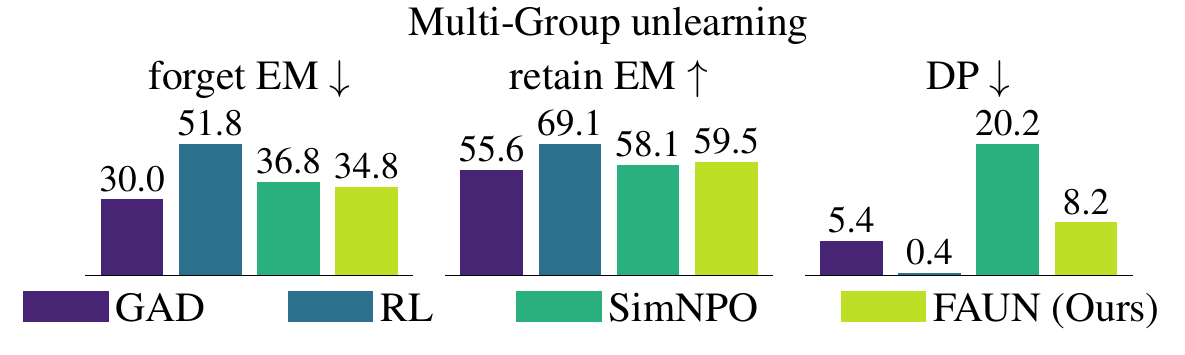}};
        \begin{scope}[x={(image.south east)},y={(image.north west)}]
            \node[font=\fontsize{6}{7}\selectfont] at (0.2,0.132) {\cite{liu2022continual}};
            \node[font=\fontsize{6}{7}\selectfont] at (0.394,0.132) {\cite{Golatkar_2020_CVPR}};
            \node[font=\fontsize{6}{7}\selectfont] at (0.666,0.132) {\cite{fan2024simplicity}};
        \end{scope}
    \end{tikzpicture}
    \vspace{-4pt}
        \caption{
        \textbf{Multi-group unlearning on \qwen}. Performance of methods where the forget set comprises 75\% \protectedattr{man} with \targetattr{right-wing} political orientation, and 25\% is sampled uniformly.
        }
        \label{fig:mixed}
    \end{minipage}
    
\end{figure}

%% file: sec/7_conclusion.tex
\section{Conclusions}

In this work, we revisit multimodal machine unlearning through the lens of fairness.
To evaluate multimodal unlearning under unbalanced unlearning scenarios, we introduce \ourbenchmark{}, a large-scale fictitious-identities benchmark, in which forget requests are biased toward a single demographic group.
To address this setting, we propose \ourmethod{}, a fairness-aware unlearning strategy that uses permanent activation steering with bias-informed PCA.
Across extensive experiments, \ourmethod{} delivers strong forgetting while substantially reducing demographic bias, highlighting overlooked fairness risks and offering a first step toward socially responsible, regulation-aligned multimodal unlearning.

%% file: sec/A_data_generation.tex
\appendix
\clearpage
\setcounter{page}{1}

\begin{center}
    \Large 
    \textbf{Unlearning Under Imbalance: Benchmarking Fairness in Multimodal LLM Unlearning} \\
    \vspace{5pt}
    Supplementary Material
\end{center}

This Appendix provides additional details on \ourbenchmark{} data generation process (\cref{appx:data_generation}), the used metrics (\cref{appx:metrics}), as well as implementation details of the approach (\cref{sec:appx:implementation-details}), additional results (\cref{sec:appx:additional-results}), and further information about the naive PCA baseline for debiasing (\cref{appx:naive_pca}). 

\section{Additional data generation details}
\label{appx:data_generation}
This section provides additional details on the data generation pipeline of \ourbenchmark{}. 
Specifically, \cref{appx:data_generation:stratified_sampling} reports the stratified sampling strategy employed for balancing visual attributes, while \cref{appx:data_generation:attributes_assignment} details how textual attributes are assigned to identities. 

\subsection{Visual attributes sampling}
\label{appx:data_generation:stratified_sampling}
We generate fictitious identities using the well-established StyleGAN2~\cite{karras2019style}, which offers realistic and diversified profile pictures with limited computational requirements. Nonetheless, StyleGAN2 has notable limitations: it cannot condition image generation on specific facial attributes, and because it is trained on FFHQ~\cite{8977347}, its outputs inherit the demographic imbalances of the underlying dataset. To mitigate these issues, we first sample a large pool of over 450k generated identities, annotate them with age, gender, and ethnicity relying on FairFace~\cite{karkkainen2021fairface}, and only then perform stratified sampling to select a final subset of 4k identities. To this end, we exploit the annotated attributes and create a strata for each unique \textit{(age group, gender, ethnicity)} triplet. Hence, identities are uniformly sampled from each stratum until the desired size of the split is reached. This guarantees that each combination of age group, gender, and ethnicity is equally represented, thereby mitigating potential visual biases in the dataset. In \cref{fig:three_stacked} we visualize the images associated to three identities.

\subsection{Textual attributes assignment}
\label{appx:data_generation:attributes_assignment}
Starting from the visual attributes provided by identities images, we construct rich identity profiles annotating other 10 sensitive attributes. 
To generate them, we noted that relying on LLM-generated textual attributes (\eg, name, education, employment) can lead to biased, inconsistent, or repetitive outputs (\eg, prompting GPT-4o to generate attributes for the ``residence'' field often results in the same city being assigned to multiple individuals). As a consequence, models will exploit these patterns and learn shortcuts that lead to overoptimistic evaluations where high accuracy is achieved by ``guessing'' the most frequent attributes rather than memorization, even on unseen identities; thus, we design a rule-based system to generate textual attributes that are consistent with the visual ones (\cref{appx:data_generation:stratified_sampling}). For this reason, some attributes should be correlated with the visual features of the individuals (\eg, education with age, birthplace with ethnicity), while others should be randomly selected from a pool of possible values (\eg, political leaning, relationship status, job). 

We map each ethnicity to a macro-area (\eg, East Asia, South Asia, Europe, etc.) based on population statistics. Next, we select a country from the chosen macro-area and generate birthplace and residence by sampling from the 10 most populous cities in that country. Names are generated by randomly combining first and last names typical of the selected country, using country-specific name templates.
The attributes for height and relationship status are determined based on age ranges, while political leaning is randomly assigned with equal probability for individuals over 14 years old.
Education and employment are assigned based on the age group annotation. Education is set according to age ranges: individuals under 7 years old have no education, those between 7 and 18 are assigned a random school name (elementary or high school), and those over 18 are assigned a university name from the country of residence or from a set of common study destinations with a certain probability.
Employment is only assigned to individuals over 18 y.o.\ by randomly selecting a job title from a predefined list, with salaries determined based on typical earnings for that job. Following the prompt in \cref{fig:template_prompt} we generate attribute specialized Q\&A templates (\cref{fig:template_sample}). Final training/unlearning samples are generated by instantiating templates with actual identity information (\cref{fig:qa-samples}).

%% file: sec/E_metrics.tex
\section{Metrics}
\label{appx:metrics}
This section provides further details about metrics used in the main paper, namely forget and retain exact match, forget and retain normalized probabilities, membership inference attack, and demographic parity.

\paragraph{Exact match (EM)}~\cite{mabenchmarking} evaluates MLLMs' accuracy in correctly predicting target attributes.
Let $(\sample, \target)$ be an example.
We measure EM by sampling a model answer $\predanswer\sim\model(\cdot\mid\sample)$, and check whether the correct target attribute $\target$ is contained in $\predanswer$.
For instance, given user query $\sample=$``Where does John live?'' and target attribute $\target=$``New York'', the sampled answer $\predanswer=$``John lives in New York.'' is correct.
Ideally, this metric should be low for the forget set and high for the retain set.

\paragraph{Normalized probability (prob.).}
The exact match can mark an answer correct when it is not (\eg, ``not New York'') and incorrect when it actually is (\eg, NY vs. New York).
To mitigate EM limitations, we also evaluate the probability of outputting the correct answer~\cite{mainitofu}.
Formally, let $\answer$ be a ground truth answer for user query $\sample$, which contains textual attribute $\target$.
Then, the normalized probability is computed as:
\begin{equation}
    \text{prob.} = p(\answer|\sample;\theta)^{1/|\answer|},
\end{equation}
where $\theta$ are model weights.
As for EM, the probability of generating the ground truth answer should be low for the forget set and high for the retain one.
Although the normalized probability is robust to negations and paraphrasing, it is vulnerable to \textit{signal dilution}, \ie, as the proportion of target-related tokens decreases relative to the total answer length, the metric becomes less sensitive to changes in unlearning effectiveness.

\paragraph{Membership inference attack (MIA).}
As \cref{subsec:eval_pipeline} describes, we follow existing works in machine unlearning~\cite{fan2023salun,mainitofu} and also compute the membership inference attack via the SOTA Min-K++~\cite{zhang2025min}.
The Min-K++ assigns a score to the sentence based on the top-$k$ tokens with the lowest normalized log probability.
Thus, the metric is defined as:
\begin{align}
    \text{Min-K++}_\text{token}(\answer[t]) &= \frac{\log p(\answer[t]|\sample,\answer[<t]) - \mu_{\answer[<t]}}{\sigma_{\answer[<t]}},\\
    \text{Min-K++}(\answer) &= \frac1{|\text{min-}k\%|}\sum_{\answer[t]\in\text{min-}k\%}\text{Min-K++}_\text{token}(\answer[t]),
\end{align}
where $\answer[t]$ is the $t$-th generated token, $\frac1{|\text{min-}k\%|}$ is the set of tokens with minimum score, $\mu_{\answer[<t]} = \mathbb{E}_{z\sim p (\cdot|\sample,\answer[<t];\theta)}[\log p(z|\sample,\answer[<t]; \theta)]$ is the expectation of the log prob.\ over the model's vocabulary, and $\sigma_{\answer[<t]} = \sqrt{\mathbb{E}_{z\sim p (\cdot|\sample,\answer[<t];\theta)}[(\log p(z|\sample,\answer[<t]; \theta) - \mu_{\answer[<t]})^2]}$ is the standard deviation.
Tokens with low probability (\ie, low score) are less likely to appear in training samples; therefore, high Min-K++ values correlate with training set membership.
To more comprehensively assess unlearning effectiveness, we compute the AUC-ROC between the scores of the forget and retain samples, where higher values indicate stronger unlearning.
We refer to the original paper for further details~\cite{zhang2025min}.

\paragraph{Demographic parity (DP)} quantifies model fairness as the difference in probability of predicting the positive class (\ie, the textual attribute) regardless of the protected attribute (\ie, the visual attribute).
Formally, the DP is defined as:
\begin{equation}
    \text{DP} = \left| P(\hat{Y} = y \mid A=\attribute) - P(\hat{Y} = y \mid A=\neg\attribute) \right|,
\end{equation}
with $y$ the target attribute. Consider, for instance, the following case: ``old men are richer, on average, than other people.''
To test whether the model is biased towards predicting that old men are richer, we compute the absolute probability difference in predicting ``\targetattr{rich}'' given ``\protectedattr{old man}'' and given ``not \protectedattr{old man}.''

Computing this metric requires an unseen set because the unlearning split contains the same identities in both training and test sets. 
Calculating DP on the unlearning split would fail to reveal fairness issues, as the model has memorized the training samples and would likely assign similar probabilities to all groups, even if biased.
Therefore, DP is computed on a held-out set, referred to as the ``fairness split'' in the main paper.
Since the model has no prior exposure to these identities, it must rely on its internal statistics and the visual attributes of each identity to make predictions. 
Consequently, if the model is biased, it might classify, for example, ``\protectedattr{old man}'' as rich more often, leading to DP $> 0$.

%% file: sec/C_implementation_details.tex
\section{Implementation details}
\label{sec:appx:implementation-details}

\subsection{Hyperparameter settings}

All fine-tuning and unlearning experiments are conducted on 4 Nvidia Ampere A100 GPUs with 64GB of memory. Both \qwen and \idefics are trained with LoRA~\cite{hu2022lowrank} with rank 64 and alpha 128. 

\paragraph{Fine-tuning.} The learning rate is set to $8\times10^{-5}$ with batch size of 16. Fine-tuning is regularized with validation samples from VQAv2~\cite{goyal2017making} (15\% of forget size) to preserve model utility.

\paragraph{Unlearning.} During unlearning, we set the batch size to 8, while we searched for the best lr within $[1\times10^{-7},1\times10^{-3}]$, which depends on the unlearning method and the forget set size.
In order to keep the comparison fair between unlearning methods, we fix the compute budget and run each method with different hyperparameter configurations for the same wall-clock time; the best run is then selected for the final comparison. 
In our setting, we leverage GAD as the budget reference as it performs one epoch over the forget set and one epoch over a subset of the retain set, corresponding to twice the size of the forget set. 
Thus, unlearning accounts for about 4\%-15\% of the fine-tuning budget, depending on the forget set size.
On a computational side, \ourmethod{} has similar requirements to the compared approaches. PCA is cheap ($\sim$35s using forget + subset of retain), while activation steering takes $\sim$2m.
\input{algorithms/bias_informed_pca}
\input{algorithms/activation_steering}
\subsection{Pseudocodes}
To further promote reproducibility, \cref{algo:pca,algo:act_steering} report PyTorch-like pseudocodes, showing how to de-bias forget set activations.

\Cref{algo:pca} shows how to compute our bias-informed PCA starting from retain and forget set activations, and the number of components to preserve $\topcomponents$.
First, activations are concatenated to retrieve an estimate of the entire training set activations, which can be used to compute the PCA via \smalltt{torch.linalg.eig}.
Then, we project the average forget set activation onto the eigenvector space \smalltt{V}.
Finally, we compute the worst-$\topcomponents$ components \wrt the projected forget activation, which correspond to orthonormal vectors that are not related to the group information; thus, we return the subspace \smalltt{V\_c} (\ie, $\components[\topcomponents]$), and the centering vector \smalltt{avg\_train\_act} (\ie, $\avghidden[tr]$).

\Cref{algo:act_steering}, instead, shows how to compute the unbiased activation steering vector for a forget identity, using subspace \smalltt{V\_c}.
First, we compute the unbiased projection for each activation independently following \cref{eq:project_unbias}.
Then, de-biased activations are averaged to obtain the unbiased retain activation \smalltt{unb\_retain\_act} (\ie, $\unbiasedhidden[r]$) and forget id activation \smalltt{unb\_forget\_id\_act} (\ie, $\unbiasedhidden[f_i]$).
Finally, the difference between these two terms yields the unbiased steering vector for a specific identity \smalltt{u\_unb\_id} (\ie, $\unbiasedsteering[f_i]$).
Note that \cref{algo:act_steering} illustrates how to compute the unbiased steering vector for a single identity for clarity. Extending this to multiple forget identities is straightforward: one can simply include an additional vector indicating the identity associated with each activation, enabling the averages to be computed correctly.

%% file: algorithms/bias_informed_pca.tex
\begin{algorithm}
\begin{lstlisting}
def bias_informed_pca(
    retain_acts: Tensor,    # [n_retain_samples, dim]
    forget_acts: Tensor,    # [n_forget_samples, dim]
    c: int,
) -> Tuple[Tensor, Tensor]: # ([dim, c], [dim])
    # compute pca
    train_acts = torch.cat((retain_acts, forget_acts))
    avg_train_act = train_acts.mean(dim=0)
    cent_train_act = train_acts - avg_train_act
    L, V = torch.linalg.eig(cent_train_act)
  
    # find directions related to group information
    avg_forget_act = forget_acts.mean(dim=0)
    cent_forget_act = avg_forget_act - avg_train_act
    proj_forget_act = V.T @ cent_forget_act
  
    # remove components
    mask = torch.topk(
        proj_forget_acts, 
        k=c, 
        largest=False
    )
    V_c = V[:,mask]
    return V_c, avg_train_act
\end{lstlisting}
\caption{Bias-informed PCA pseudocode}
\label{algo:pca}
\end{algorithm}

%% file: algorithms/activation_steering.tex
\begin{algorithm}
\begin{lstlisting}
def compute_steering_vector(
    retain_acts: Tensor,    # [n_retain_samples, dim]
    forget_id_acts: Tensor, # [n_forget_id_samples, dim]
    V_c: Tensor,            # [dim, c]
    avg_train_act: Tensor,  # [dim]
) -> Tensor:                # [dim]
    # project in the unbiased subspace
    unb_retain_acts = (
        V_c @ V_c.T @ (retain_acts - avg_train_act)
    ) + avg_train_act
    unb_forget_id_acts = (
        V_c @ V_c.T @ (forget_id_acts - avg_train_act)
    ) + avg_train_act

    # compute avg activation
    unb_retain_act = unb_retain_acts.mean(dim=0)
    unb_forget_id_act = unb_forget_id_acts.mean(dim=0)

    # compute steering vector
    u_unb_id = unb_forget_id_act - unb_retain_act
    return u_unb_id
\end{lstlisting}
\caption{Unbiased activation steering vector}
\label{algo:act_steering}
\end{algorithm}

%% file: sec/D_additional_results.tex
\section{Additional results}
\label{sec:appx:additional-results}
\input{tables/qwen_additional}
\input{tables/idefics_additional}
\input{tables/qwen_105090}
\input{tables/qwen_bert}
\input{tables/idefics_bert}

\Cref{tab:qwen_supp,tab:idefics_supp} report additional results for \targetattr{single}-\protectedattr{asians} and \targetattr{tall}-\protectedattr{females} settings on \qwen and \idefics MLLMs, respectively.
As can be noted, \ourmethod{} confirms a good tradeoff between unlearning quality and fairness. In all settings, \ourmethod{} achieves the best or second best forgetting, as demonstrated by the EM metric (\eg, 40.49\% EM \vs 46.49\% of SimNPO), while preserving its utility on retain data (\eg 59.98\% \vs 58.33\% of SimNPO for the same setting). The consistently high MIA metric further shows that retain and forget samples become distinguishable from a statistical perspective, with a slight drop compared to SimNPO in the single-asian setting for \qwen. When looking at the demographic parity, \ourmethod{} achieves the lowest DP among methods with similar unlearning quality in two out of four settings, while yielding competitive performance with the baselines in the other two.

In \cref{tab:qwen_supp_ratio} we experiment with increasing unlearning ratios from 10\% to 90\% of \targetattr{single}-\protectedattr{asians}. At 10\%, \ourmethod{} achieves solid forgetting (42.11\% EM) while preserving retain EM at 54.89\% with a low DP (3.65), substantially improving over baseline methods (\eg, GAD 21.41 DP and RL 21.50 DP). In the more challenging 50\% ratio, \ourmethod{} yields the best unlearning quality while maintaining competitive fairness (5.38 DP \vs 4.54 of SimNPO). Even under the extreme 90\% ratio, where most baselines collapse (\eg, GA drops to 43.51\% retain EM, GAD reaches DP = 11.79), FAUN sustains both high retain accuracy (62.70\% EM) and moderate fairness (DP = 4.80). Looking at the avg.~gap, \ourmethod{} and SimNPO yield the best performance with baselines failing in at least one setting. \ourmethod{} delivers good forgetting, strong retain utility, and consistently controlled demographic disparity across all unlearning regimes.

\cref{tab:qwen_bert} and \cref{tab:idefics_bert} complement the main paper results by reporting ROUGE-L~\cite{lin2004rouge} and BERTScore~\cite{zhangbertscore} metrics, both computed between predicted and ground-truth answers.
Unlike exact-match (EM) metrics, these measures are not constrained by strict string matching and can capture more nuanced similarities between answers.
However, since our benchmark is designed to evaluate the model's knowledge on memorized retain and forget information, questions are direct, and answers are usually short with low entropy (\eg, ``John Doe is born in New York'').
For this reason, EM on the target attribute provides a reliable estimate of unlearning effectiveness and remains sensitive to the quality of the unlearning procedure (\eg, 74.57 for the original model vs.\ 17.05 for the retrained model on forget EM).
In contrast, we empirically observe that ROUGE-L and BERTScore exhibit limited sensitivity to the unlearning approach (\eg, ROUGE-L: 77.09 \vs 66.55 and BERTScore: 93.89 \vs 90.04 between the original and retrained models), making them less effective for comparing unlearning performance across models.
Nevertheless, results on both the retain and forget sets are \textit{consistent} with the results shown in the main paper.

\begin{figure}
    \centering
    \begin{subfigure}{\textwidth}
        \centering
        \includegraphics[width=0.9\linewidth]{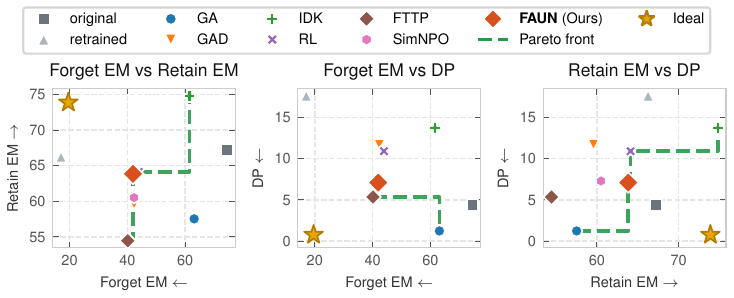}
        \caption{Unlearning \protectedattr{men} with \targetattr{right-wing} political orientation on \qwen{}.}
        \label{fig:pareto_qwen_male_right}
    \end{subfigure}
    \begin{subfigure}{\textwidth}
        \centering
        \includegraphics[width=0.9\linewidth]{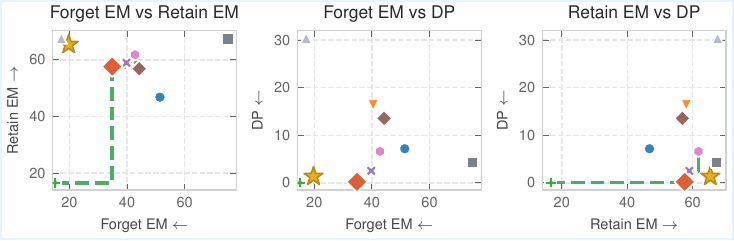}
        \caption{Unlearning \protectedattr{adults} with \targetattr{low} net-worth on \qwen{}.}
        \label{fig:pareto_qwen_adult_lownw}
    \end{subfigure}
    \begin{subfigure}{\textwidth}
        \centering
        \includegraphics[width=0.9\linewidth]{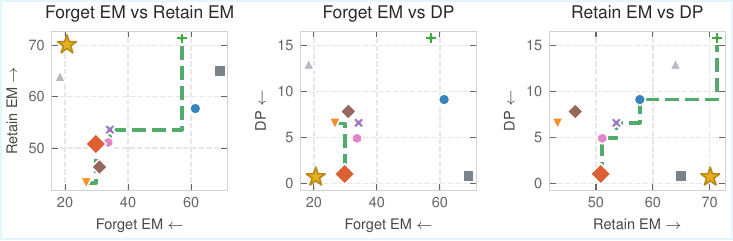}
        \caption{Unlearning \protectedattr{men} with \targetattr{right-wing} political orientation on \idefics{}.}
        \label{fig:pareto_idefics_male_right}
    \end{subfigure}
    \begin{subfigure}{\textwidth}
        \centering
        \includegraphics[width=0.9\linewidth]{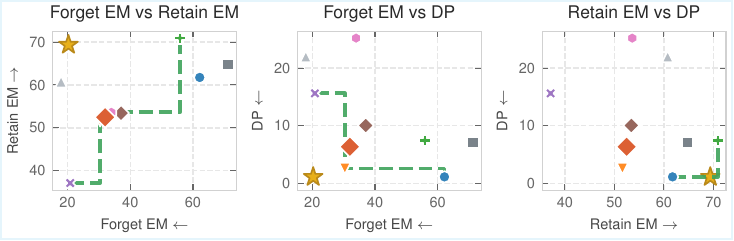}
        \caption{Unlearning \protectedattr{adults} with \targetattr{low} net-worth on \idefics{}.}
        \label{fig:pareto_idefics_adult_lownw}
    \end{subfigure}
    \caption{\textbf{Pareto frontier comparisons of unlearning algorithms.} Points show methods evaluated on Forget EM vs Retain EM (left), Forget EM vs DP (center), and Retain EM vs DP (right). Dashed lines indicate the Pareto frontier, and the star the ideal result.}
    \label{fig:pareto_all}
\end{figure}
Finally, \cref{fig:pareto_all} visualizes the trade-offs among competing objectives in unbalanced machine unlearning, by reporting their Pareto frontiers.
Each subplot (\cref{fig:pareto_qwen_male_right,fig:pareto_qwen_adult_lownw,fig:pareto_idefics_male_right,fig:pareto_idefics_adult_lownw}) compares two objectives (forget EM \vs retain EM, forget EM \vs DP, and retain EM \vs DP) with models positioned according to their performance along the corresponding pair of axes.
Compared to existing baselines, \ourmethod{} consistently achieves a more favorable trade-off between forget EM, retain EM, and DP.
For instance, in \cref{fig:pareto_qwen_male_right}, our method lies on the Pareto frontier in two out of the three comparisons.
In the remaining comparison, although our method does not lie exactly on the frontier, its performance remains very close to the frontier and comparable to methods that do appear on it.
Importantly, the methods that dominate the frontier in that particular case, FTTP and GA, do not perform consistently across the other comparisons.
Similar trends appear in the Idefics3-8B experiments (\cref{fig:pareto_idefics_male_right,fig:pareto_idefics_adult_lownw}), where our method generally lies closer to the Pareto frontier than several baselines, consistently providing low DP compared to GA and GAD while maintaining stronger retention than methods that aggressively forget.

\subsection{Sensitivity to number of components}
\begin{figure}[!ht]
    \centering
\includegraphics[width=\linewidth]{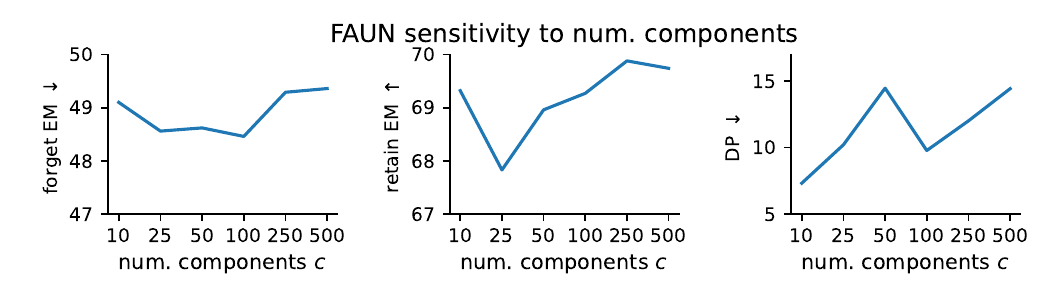}
\caption{Sensitivity of \ourmethod{} to the number of components $c$. The ablation is computed on \qwen{}, by unlearning 50\% of \protectedattr{man} with \targetattr{right-wing} political orientation.}
\label{fig:sensitivity-components}
\end{figure}

In \cref{fig:sensitivity-components} we report the unlearning and fairness performance when ablating on the number of components $c$ that are preserved for the unbiased subspace. As can be noted, EM is stable at different numbers of components $c$, DP increases as expected. 

%% file: tables/qwen_additional.tex
\begin{table}[t]
    
    \centering
    \caption{\textbf{Unbalanced unlearning on \qwen.} Unlearning performance of different methods where the forget set is composed of 50\% of all datapoints of a single group.}
    \resizebox{0.8\linewidth}{!}{%
    \scriptsize
    \begin{tabularx}{\linewidth}{
        L{1.8cm}
        *{7}Y
    }
    \multirow{2.5}{*}{method} & \multicolumn{2}{c}{forget} & \multicolumn{2}{c}{retain} & \multirow{2.5}{*}{MIA$\uparrow$} & \multirow{2.5}{*}{DP$\downarrow$} & \multirow{2.5}{*}{avg.~gap $\downarrow$} \\
\cmidrule(lr){2-3}
\cmidrule(lr){4-5}
    & EM~$\downarrow$ & prob.~$\downarrow$ & EM~$\uparrow$ & prob.~$\uparrow$ & \\
\toprule
    \multicolumn{8}{l}{\textcolor{gray}{\textit{\targetattr{single} \protectedattr{asians}}}}\\
    original & 67.76 & 46.70 & 67.98 & 49.30 & 50.30 & 4.64 & \cellcolor{experiment}- \\
    retrained & 20.51 & 29.30 & 65.85 & 49.00 & 61.80 & 6.36 & \cellcolor{experiment}- \\
\cmidrule(lr){1-8}
\textcolor{gray}{MIU~\cite{de2025group}} & \textcolor{gray}{12.64} & \textcolor{gray}{26.70} & \textcolor{gray}{68.58} & \textcolor{gray}{44.53} & \textcolor{gray}{35.13} & \textcolor{gray}{14.19} &	\cellcolor{experiment}{\textcolor{gray}{9.41}} \\
\hdashline
    GA~\cite{thudi2022unrolling} & 53.67 & 43.50 & 58.60 & 45.40 & 49.90 & 8.01 & \cellcolor{experiment}13.02 \\
    GAD~\cite{liu2022continual} & 46.84 & 41.00 & 57.22 & 45.90 & 52.60 & 4.61 & \cellcolor{experiment}10.59 \\
    IDK~\cite{mainitofu} & 56.84 & 44.80 & 65.99 & 49.20 & 51.90 & 4.74 & \cellcolor{experiment}11.13 \\
    RL~\cite{Golatkar_2020_CVPR} & 41.12 & 41.60 & 54.65 & 46.80 & 52.40 & 4.13 & \cellcolor{experiment}9.97 \\
    SimNPO~\cite{fan2024simplicity} & 43.57 & 41.70 & 57.47 & 48.20 & 53.80 & 4.54 & \cellcolor{experiment}\underline{9.53} \\
    LUNAR~\cite{shen2025lunar} & 13.42 & 26.97 & 16.35 & 28.89 & 49.95 & 5.39 &	\cellcolor{experiment}{16.27} \\
    \textbf{\ourmethod{}}~\textbf{(Ours)} & 41.43 & 41.80 & 58.48 & 47.90 & 52.70 & 5.38 & \cellcolor{experiment}\textbf{9.39} \\ [0.5em]
    
    \multicolumn{7}{l}{\textcolor{gray}{\textit{\targetattr{tall} \protectedattr{females}}}}\\
    original & 68.02 &	46.90 &	67.96 &	49.20 &	48.20 &	6.99 & \cellcolor{experiment}- \\
    retrained & 16.91 &	28.20 &	66.67 &	49.20 &	61.40 &	14.89 & \cellcolor{experiment}- \\
\cmidrule(lr){1-8}
\textcolor{gray}{MIU~\cite{de2025group}} & \textcolor{gray}{42.36} & \textcolor{gray}{29.76} & \textcolor{gray}{67.98} & \textcolor{gray}{40.40} & \textcolor{gray}{33.25} & \textcolor{gray}{0.79} &	\cellcolor{experiment}{\textcolor{gray}{13.22}} \\
\hdashline
    GA~\cite{thudi2022unrolling} & 63.83 & 44.50 & 62.27 & 45.90 & 47.90 & 5.64 & \cellcolor{experiment}15.01 \\
    GAD~\cite{liu2022continual} & 50.00 & 43.10 & 64.47 & 48.20 & 50.20 & 13.30 & \cellcolor{experiment}12.62 \\
    IDK~\cite{mainitofu} & 52.47 & 43.20 & 65.03 & 48.20 & 50.00 & 1.88 & \cellcolor{experiment}\underline{11.08} \\
    RL~\cite{Golatkar_2020_CVPR} & 48.27 & 43.40 & 61.64 & 48.40 & 50.20 & 6.54 & \cellcolor{experiment}11.69 \\
    SimNPO~\cite{fan2024simplicity} & 46.42 & 42.30 & 58.33 & 48.30 & 52.80 & 7.01 & \cellcolor{experiment}11.41 \\
    LUNAR~\cite{shen2025lunar} & 14.91 & 29.07 & 18.00 & 30.51 & 48.63 & 3.09 &	\cellcolor{experiment}{15.80} \\
    \textbf{\ourmethod{}}~\textbf{(Ours)} & 40.49 & 40.90 & 59.98 & 47.60 & 52.80 & 3.33 & \cellcolor{experiment}\textbf{10.78} \\
    \end{tabularx}
    }
    \label{tab:qwen_supp}
\end{table}

%% file: tables/idefics_additional.tex
\begin{table}[t]
    
    \centering
    \caption{\textbf{Unbalanced unlearning on \idefics.} Unlearning performance of different methods where the forget set is composed of 50\% of all datapoints of a single group.}
    \resizebox{0.8\linewidth}{!}{%
    \scriptsize
    \begin{tabularx}{\linewidth}{
        L{1.8cm}
        *{7}Y
    }
    \multirow{2.5}{*}{method} & \multicolumn{2}{c}{forget} & \multicolumn{2}{c}{retain} & \multirow{2.5}{*}{MIA$\uparrow$} & \multirow{2.5}{*}{DP$\downarrow$} & \multirow{2.5}{*}{avg.~gap $\downarrow$} \\
\cmidrule(lr){2-3}
\cmidrule(lr){4-5}
    & EM~$\downarrow$ & prob.~$\downarrow$ & EM~$\uparrow$ & prob.~$\uparrow$ & \\
\toprule
    \multicolumn{7}{l}{\textcolor{gray}{\textit{\targetattr{single} \protectedattr{asians}}}}\\
    original & 72.96 &	47.50 &	64.92 &	48.50 &	48.80 &	9.08 & \cellcolor{experiment}- \\
    retrained & 16.02 &	28.30 &	63.19 &	47.30 &	62.00 &	8.18 & \cellcolor{experiment}- \\
\cmidrule(lr){1-8}
\textcolor{gray}{MIU~\cite{de2025group}} & \textcolor{gray}{21.27} & \textcolor{gray}{29.90} & \textcolor{gray}{62.29} & \textcolor{gray}{39.26} & \textcolor{gray}{30.80} & \textcolor{gray}{6.12} &\cellcolor{experiment}{\textcolor{gray}{8.84}} \\
\hdashline
    GA~\cite{thudi2022unrolling} & 67.96 & 45.40 & 60.04 & 45.30 & 47.50 & 6.94 & \cellcolor{experiment}15.94 \\
    GAD~\cite{liu2022continual} & 46.33 & 39.60 & 54.40 & 42.70 & 49.60 & 5.96 & \cellcolor{experiment}12.23 \\
    IDK~\cite{mainitofu} & 43.88 & 41.40 & 54.86 & 44.70 & 50.50 & 7.73 & \cellcolor{experiment}11.85 \\
    RL~\cite{Golatkar_2020_CVPR} & 35.82 & 39.20 & 50.86 & 44.60 & 52.50 & 9.20 & \cellcolor{experiment}\underline{10.74} \\
    SimNPO~\cite{fan2024simplicity} & 38.67 & 39.50 & 50.81 & 45.40 & 52.10 & 9.83 & \cellcolor{experiment}11.31 \\
    LUNAR~\cite{shen2025lunar} & 64.60 & 42.68 & 63.37 & 43.10 & 49.11 & 8.84 &	\cellcolor{experiment}{14.84} \\
    \textbf{\ourmethod{}}~\textbf{(Ours)} & 37.55 & 36.70 & 51.03 & 42.50 & 52.60 & 5.72 & \cellcolor{experiment}\textbf{10.33} \\ [0.5em]
    
    \multicolumn{7}{l}{\textcolor{gray}{\textit{\targetattr{tall} \protectedattr{females}}}}\\
    original & 64.07 &	45.90 &	65.45 &	48.60 &	49.80 &	1.43 & \cellcolor{experiment}- \\
    retrained & 18.15 &	28.30 &	63.50 &	47.50 &	62.60 &	22.46 & \cellcolor{experiment}- \\
\cmidrule(lr){1-8}
\textcolor{gray}{MIU~\cite{de2025group}} & \textcolor{gray}{28.02} & \textcolor{gray}{44.51} & \textcolor{gray}{58.75} & \textcolor{gray}{51.01} & \textcolor{gray}{37.19} & \textcolor{gray}{2.05} &	\cellcolor{experiment}{\textcolor{gray}{10.30}} \\
\hdashline
    GA~\cite{thudi2022unrolling} & 58.89 & 43.10 & 60.56 & 45.40 & 49.40 & 0.12 & \cellcolor{experiment}12.32 \\
    GAD~\cite{liu2022continual} & 40.99 & 40.60 & 54.92 & 45.10 & 50.40 & 6.57 & \cellcolor{experiment}10.82 \\
    IDK~\cite{mainitofu} & 41.48 & 41.50 & 57.50 & 46.30 & 50.90 & 12.53 & \cellcolor{experiment}11.33 \\
    RL~\cite{Golatkar_2020_CVPR} & 33.21 & 39.90 & 51.68 & 45.60 & 50.30 & 3.44 & \cellcolor{experiment}9.35 \\
    SimNPO~\cite{fan2024simplicity} & 31.85 & 35.50 & 48.90 & 42.80 & 52.30 & 0.29 & \cellcolor{experiment}\textbf{8.47} \\
    LUNAR~\cite{shen2025lunar} & 59.92 & 41.74 & 63.48 & 43.21 & 49.96 & 1.08 &	\cellcolor{experiment}{12.21} \\
    \textbf{\ourmethod{}}~\textbf{(Ours)} & 28.40 & 36.60 & 50.06 & 43.70 & 53.80 & 10.57 & \cellcolor{experiment}\underline{9.19} \\
    \end{tabularx}
    }
    \label{tab:idefics_supp}
\end{table}

%% file: tables/qwen_105090.tex
\begin{table}[t]
    \centering
    \caption{\textbf{Unbalanced unlearning on \qwen.} Unlearning performance of different methods where the forget set is composed of 10\%, 50\% and 90\% of all datapoints of a single group.}
    \resizebox{0.8\linewidth}{!}{%
    \scriptsize
    \begin{tabularx}{\linewidth}{
        L{1.8cm}
        *{6}Y
        Y
    }
    \multirow{2.5}{*}{method} & \multicolumn{2}{c}{forget} & \multicolumn{2}{c}{retain} & \multirow{2.5}{*}{MIA$\uparrow$} & \multirow{2.5}{*}{DP$\downarrow$} & \multirow{2.5}{*}{avg.~gap $\downarrow$} \\
\cmidrule(lr){2-3}
\cmidrule(lr){4-5}
    & EM~$\downarrow$ & prob.~$\downarrow$ & EM~$\uparrow$ & prob.~$\uparrow$ \\
\toprule
    \multicolumn{8}{l}{\textcolor{gray}{\textit{\targetattr{single} \protectedattr{asians} @ 10\% forget rate}}}\\
    original & 64.74 & 47.30 & 68.00 & 49.20 & 49.20 & 4.64 & \cellcolor{experiment}- \\
    retrained & 24.74 & 31.20 & 69.27 & 48.70 & 58.60 & 9.85 & \cellcolor{experiment}- \\
\cmidrule(lr){1-8}
\textcolor{gray}{MIU~\cite{de2025group}} & \textcolor{gray}{49.80} & \textcolor{gray}{35.19} & \textcolor{gray}{65.23} & \textcolor{gray}{37.99} & \textcolor{gray}{51.02} & \textcolor{gray}{5.19} & \cellcolor{experiment}{\textcolor{gray}{9.42}} \\
\hdashline
    GA~\cite{thudi2022unrolling} & 31.58 & 37.10 & 50.30 & 44.90 & 52.30 & 8.12 & \cellcolor{experiment}\textbf{8.32} \\
    GAD~\cite{liu2022continual} & 52.63 & 45.90 & 59.94 & 48.80 & 49.40 & 21.41 & \cellcolor{experiment}13.77 \\
    IDK~\cite{mainitofu} & 45.26 & 44.60 & 58.51 & 48.10 & 49.90 & 5.71 & \cellcolor{experiment}9.94 \\
    RL~\cite{Golatkar_2020_CVPR} & 48.95 & 45.80 & 62.83 & 48.70 & 48.10 & 21.50 & \cellcolor{experiment}12.87 \\
    SimNPO~\cite{fan2024simplicity} & 44.74 & 41.90 & 58.53 & 48.60 & 53.20 & 7.25 & \cellcolor{experiment}\underline{9.03} \\
    LUNAR~\cite{shen2025lunar} & 14.98 & 29.33 & 20.05 & 32.66 & 49.51 & 7.10 & \cellcolor{experiment}{15.51} \\
    \textbf{\ourmethod{}}~\textbf{(Ours)} & 42.11 & 44.90 & 54.89 & 47.90 & 49.30 & 3.65 & \cellcolor{experiment}9.86 \\ [0.5em]
    \multicolumn{8}{l}{\textcolor{gray}{\textit{\targetattr{single} \protectedattr{asians} @ 50\% forget rate}}}\\
    original & 67.76 & 46.70 & 67.98 & 49.30 & 50.30 & 4.64 & \cellcolor{experiment}- \\
    retrained & 20.51 & 29.30 & 65.85 & 49.00 & 61.80 & 6.36 & \cellcolor{experiment}- \\
\cmidrule(lr){1-8}
\textcolor{gray}{MIU~\cite{de2025group}} & \textcolor{gray}{12.64} & \textcolor{gray}{26.70} & \textcolor{gray}{68.58} & \textcolor{gray}{44.53} & \textcolor{gray}{35.13} & \textcolor{gray}{14.19} &	\cellcolor{experiment}{\textcolor{gray}{9.41}} \\
\hdashline
    GA~\cite{thudi2022unrolling} & 53.67 & 43.50 & 58.60 & 45.40 & 49.90 & 8.01 & \cellcolor{experiment}13.02 \\
    GAD~\cite{liu2022continual} & 46.84 & 41.00 & 57.22 & 45.90 & 52.60 & 4.61 & \cellcolor{experiment}10.59 \\
    IDK~\cite{mainitofu} & 56.84 & 44.80 & 65.99 & 49.20 & 51.90 & 4.74 & \cellcolor{experiment}11.13 \\
    RL~\cite{Golatkar_2020_CVPR} & 41.12 & 41.60 & 54.65 & 46.80 & 52.40 & 4.13 & \cellcolor{experiment}9.97 \\
    SimNPO~\cite{fan2024simplicity} & 43.57 & 41.70 & 57.47 & 48.20 & 53.80 & 4.54 & \cellcolor{experiment}\underline{9.53} \\
    LUNAR~\cite{shen2025lunar} & 13.42 & 26.97 & 16.35 & 28.89 & 49.95 & 5.39 &	\cellcolor{experiment}{16.04} \\
    \textbf{\ourmethod{}}~\textbf{(Ours)} & 41.43 & 41.80 & 58.48 & 47.90 & 52.70 & 5.38 & \cellcolor{experiment}\textbf{9.39} \\ [0.5em]
    \multicolumn{8}{l}{\textcolor{gray}{\textit{\targetattr{single} \protectedattr{asians} @ 90\% forget rate}}}\\
    original & 67.84 & 46.90 & 67.98 & 49.40 & 50.10 & 4.64 & \cellcolor{experiment}- \\
    retrained & 17.16 & 29.30 & 69.15 & 49.80 & 63.90 & 29.59 & \cellcolor{experiment}- \\
\cmidrule(lr){1-8}
\textcolor{gray}{MIU~\cite{de2025group}} & \textcolor{gray}{18.88} & \textcolor{gray}{33.05} & \textcolor{gray}{68.07} & \textcolor{gray}{44.92} & \textcolor{gray}{32.22} & \textcolor{gray}{3.45} &	\cellcolor{experiment}{\textcolor{gray}{7.76}} \\
\hdashline
    GA~\cite{thudi2022unrolling} & 31.65 & 36.40 & 43.51 & 43.00 & 53.20 & 9.89 & \cellcolor{experiment}12.43 \\
    GAD~\cite{liu2022continual} & 43.35 & 43.10 & 64.11 & 49.20 & 52.50 & 11.79 & \cellcolor{experiment}11.47 \\
    IDK~\cite{mainitofu} & 35.68 & 40.80 & 58.20 & 48.00 & 54.40 & 3.26 & \cellcolor{experiment}\underline{9.25} \\
    RL~\cite{Golatkar_2020_CVPR} & 44.55 & 42.40 & 63.76 & 48.60 & 52.60 & 8.50 & \cellcolor{experiment}11.14 \\
    SimNPO~\cite{fan2024simplicity} & 36.70 & 40.40 & 61.30 & 49.30 & 55.90 & 5.54 & \cellcolor{experiment}\textbf{8.75} \\
    LUNAR~\cite{shen2025lunar} & 13.42 & 26.97 & 16.35 & 28.89 & 49.95 & 5.39 & \cellcolor{experiment}{16.52} \\
    \textbf{\ourmethod{}}~\textbf{(Ours)} & 39.66 & 42.40 & 62.70 & 49.80 & 54.80 & 4.80 & \cellcolor{experiment}9.32 \\ [0.5em]
    \end{tabularx}
    }
    \label{tab:qwen_supp_ratio}
\end{table}

%% file: tables/qwen_bert.tex
\begin{table}[t]
    \centering
    \caption{\textbf{Unbalanced unlearning on \qwen.} Performance of different methods where the forget set is composed of 50\% of all datapoints of a single group.}
    \resizebox{0.8\linewidth}{!}{%
    \scriptsize
    \begin{tabularx}{\linewidth}{
        L{1.8cm}
        *{8}Y
    }

\multirow{2.5}{*}{method} & \multicolumn{3}{c}{forget} & \multicolumn{3}{c}{retain} & \multirow{2.5}{*}{DP$\downarrow$} & \multirow{2.5}{*}{avg. gap $\downarrow$} \\
\cmidrule(lr){2-4}
\cmidrule(lr){5-7}
    & EM~$\downarrow$ & ROUGE~$\downarrow$ & BERT~$\downarrow$ & EM~$\uparrow$ & ROUGE~$\uparrow$ & BERT~$\uparrow$ \\
\toprule
    \multicolumn{7}{l}{\textcolor{gray}{\textit{\protectedattr{man} with \targetattr{right-wing} political orientation}}} \\
    original & 74.57 &  77.09 & 93.89 & 67.23 & 76.05 & 93.51 & 4.35 & \cellcolor{experiment}- \\ 
    retrained & 17.05 & 66.55 & 90.04 & 66.26 & 75.93 & 93.49 & 17.65 & \cellcolor{experiment}- \\ 
\cmidrule(lr){1-9}
    \textcolor{gray}{MIU}~\cite{de2025group} & \textcolor{gray}{16.83}  & \textcolor{gray}{22.57}  & \textcolor{gray}{67.29}  & \textcolor{gray}{68.94}  & \textcolor{gray}{69.75}  & \textcolor{gray}{90.62}  & \textcolor{gray}{13.62}  & \textcolor{gray}{\cellcolor{experiment}13.19}
    \\
    \hdashline
    GA~\cite{thudi2022unrolling} & 63.13 &      74.96 & 92.9 &  57.53 & 74.27 & 92.65 & 1.24 & \cellcolor{experiment}9.97 \\ 
    GAD~\cite{liu2022continual} & 40.44 &       70.61 & 91.67 & 58.04 & 74.01 & 92.81 & 16.35 & \cellcolor{experiment}8.04 \\ 
    IDK~\cite{mainitofu} & 61.53 &      74.71 & 93.03 & 74.77 & 77.39 & 93.91 & 13.76 & \cellcolor{experiment}11.40 \\ 
    RL~\cite{Golatkar_2020_CVPR} & 39.77 &      70.90  & 91.94 & 58.96 & 74.46 & 92.95 & 2.51 & \cellcolor{experiment}\underline{5.83} \\ 
    FTTP~\cite{li2025forget} & 40.15 & 70.99 & 91.72 & 54.46& 73.74 & 92.7 & 5.36 & \cellcolor{experiment}7.05 \\ 
    SimNPO~\cite{fan2024simplicity} & 42.79 &   71.52 & 92.15 & 61.77 & 75.13 & 93.29 & 6.61 & \cellcolor{experiment}6.42 \\ 
    LUNAR~\cite{shen2025lunar} & 19.79 &   11.50 & 57.94 & 16.67 & 10.90 & 57.28 & 5.93 & \cellcolor{experiment}35.24 \\
    \textbf{\ourmethod{}}~\textbf{(Ours)}  & 34.83 &    69.52 & 91.18 & 57.59 & 74.07 & 92.72 & 0.18 & \cellcolor{experiment}\textbf{4.77} \\
    \\[-0.5em]
    
    \multicolumn{7}{l}{\textcolor{gray}{\textit{\protectedattr{adult} with \targetattr{low} net-worth}}} \\
    original & 74.91 &  76.98 & 93.89 & 67.24 & 76.06 & 93.51 & 4.25 & \cellcolor{experiment}- \\ 
    retrained & 17.20 & 65.96 & 89.83 & 67.63 & 76.14 & 93.58 & 30.42  & \cellcolor{experiment}- \\ 
\cmidrule(lr){1-9}

    \textcolor{gray}{MIU}~\cite{de2025group}  & \textcolor{gray}{22.19}  & \textcolor{gray}{43.90}  & \textcolor{gray}{78.12}  & \textcolor{gray}{68.62}  & \textcolor{gray}{70.10}  & \textcolor{gray}{90.78}  & \textcolor{gray}{11.70}  & \textcolor{gray}{\cellcolor{experiment}8.61}
    \\
    \hdashline
    GA~\cite{thudi2022unrolling} & 51.39 &      71.67 & 91.83 & 46.81 & 71.11 & 91.64 & 7.16 & \cellcolor{experiment}10.98 \\ 
    GAD~\cite{liu2022continual} & 42.31 &       71.37 & 91.95 & 59.59 & 74.72 & 93.15 & 11.70 & \cellcolor{experiment}7.75 \\ 
    IDK~\cite{mainitofu} & 15.09 &      65.92 & 89.57 & 16.63 & 65.74 & 89.4 &  0.04 & \cellcolor{experiment}9.72 \\ 
    RL~\cite{Golatkar_2020_CVPR} & 43.93 &      71.67 & 92.05 & 64.12 & 75.45 & 93.4 &  10.91 & \cellcolor{experiment}7.14 \\ 
    FTTP~\cite{li2025forget} & 44.22 & 71.66 & 91.95 & 56.86 & 74.17 & 92.89 & 13.54 & \cellcolor{experiment}8.83 \\ 
    SimNPO~\cite{fan2024simplicity} & 42.31 &   71.34 & 92.02 & 60.51 & 74.82 & 93.12 & 7.30 & \cellcolor{experiment}\underline{6.98} \\ 
    LUNAR~\cite{shen2025lunar} & 29.04  &   22.15 & 71.34 &  28.61 & 23.52 & 71.92 &  7.56 & \cellcolor{experiment}27.86 \\ 
    \textbf{\ourmethod{}}~\textbf{(Ours)} & 41.97 &     71.45 & 91.94 & 63.84 & 75.49 & 93.31 & 7.10 & \cellcolor{experiment}\textbf{6.31} \\  
    \end{tabularx}
    }
    \label{tab:qwen_bert}
\end{table}

%% file: tables/idefics_bert.tex
\begin{table}[t]
    \centering
    \caption{\textbf{Unbalanced unlearning on \idefics.} Performance of different methods where the forget set is composed of 50\% of all datapoints of a single group.}
    \resizebox{0.8\linewidth}{!}{%
    \scriptsize
    \begin{tabularx}{\linewidth}{
        L{1.8cm}
        *{8}Y
    }
\multirow{2.5}{*}{method} & \multicolumn{3}{c}{forget} & \multicolumn{3}{c}{retain} & \multirow{2.5}{*}{DP$\downarrow$} & \multirow{2.5}{*}{avg. gap $\downarrow$} \\
\cmidrule(lr){2-4}
\cmidrule(lr){5-7}
    & EM~$\downarrow$ & ROUGE~$\downarrow$ & BERT~$\downarrow$ & EM~$\uparrow$ & ROUGE~$\uparrow$ & BERT~$\uparrow$ \\
\toprule
    \multicolumn{7}{l}{\textcolor{gray}{\textit{\protectedattr{man} with \targetattr{right-wing} political orientation}}}\\
    original & 69.08 &	76.85 & 93.81 &	64.97 &	76.04 & 93.46 &	0.78 & \cellcolor{experiment}- \\
    retrained & 18.38 &	67.03 & 90.24 &	63.98 &	75.92 & 93.53 &	12.99 & \cellcolor{experiment}- \\
\cmidrule(lr){1-9}
    \textcolor{gray}{MIU}~\cite{de2025group}  & \textcolor{gray}{29.50}  & \textcolor{gray}{41.22}  & \textcolor{gray}{78.80}  & \textcolor{gray}{63.03}  & \textcolor{gray}{67.77}  & \textcolor{gray}{90.22}  & \textcolor{gray}{8.49}  & \textcolor{gray}{\cellcolor{experiment}{9.90}} \\ 
    \hdashline
    GA~\cite{thudi2022unrolling} & 61.34 &      75.76 & 93.17 & 57.71 & 75.38 & 92.97 & 9.12 & \cellcolor{experiment}10.16 \\ 
    GAD~\cite{liu2022continual} & 30.39 &       69.52 & 91.34 & 51.63 & 73.61 & 92.84 & 2.55 & \cellcolor{experiment}\underline{4.79} \\ 
    IDK~\cite{mainitofu} & 55.86 &      73.96 & 92.54 & 70.86 & 77.11 & 93.67 & 7.43 & \cellcolor{experiment}8.91 \\ 
    RL~\cite{Golatkar_2020_CVPR} & 20.84 &      68.03 & 90.95 & 37.11 & 70.92 & 91.88 & 15.61 & \cellcolor{experiment}{7.61} \\ 
    FTTP~\cite{li2025forget} & 30.95 & 70.08 & 91.39 & 46.35 & 72.67 & 92.23 &7.82 & \cellcolor{experiment}6.68 \\ 
    SimNPO~\cite{fan2024simplicity} & 33.97 &   70.31 & 91.42 & 53.60 & 74.11 & 92.77 & 25.19 & \cellcolor{experiment}{8.31} \\ 
    LUINAR~\cite{shen2025lunar} & 69.79 &   70.31 & 90.90 & 65.05 & 69.08 & 90.47 & 8.53 & \cellcolor{experiment}{10.69} \\
    \textbf{\ourmethod{}}~\textbf{(Ours)} & 29.78 &     69.40  & 91.30 &  50.83 & 73.83 & 92.82 & 1.02 & \cellcolor{experiment}\textbf{4.54} \\ 
    \\[-0.5em]
    
    \multicolumn{7}{l}{\textcolor{gray}{\textit{\protectedattr{adult} with \targetattr{low} net-worth}}} \\
    original & 71.21 &	77.17 & 93.86 &	64.78 &	76.01 & 93.45 &	7.08 & \cellcolor{experiment}- \\
    retrained & 17.94 &	67.12 & 90.2 &	60.77 &	75.33 & 93.28 &	22.00 & \cellcolor{experiment}- \\
\cmidrule(lr){1-9}
    \textcolor{gray}{MIU}~\cite{de2025group}  & \textcolor{gray}{53.80}  & \textcolor{gray}{65.10}  & \textcolor{gray}{88.90}  & \textcolor{gray}{64.98}  & \textcolor{gray}{70.25}  & \textcolor{gray}{90.87}  & \textcolor{gray}{12.80}  & \textcolor{gray}{\cellcolor{experiment}{9.10}} \\ 
    \hdashline
    GA~\cite{thudi2022unrolling} & 62.14 &      75.91 & 93.48 & 61.75 & 75.51 & 93.26 & 1.12 & \cellcolor{experiment}8.37 \\ 
    GAD~\cite{liu2022continual} & 26.71 &       68.92 & 91.12 & 43.24 & 71.84 & 92.25 & 6.52 & \cellcolor{experiment}5.72 \\ 
    IDK~\cite{mainitofu} & 57.17 &      74.62 & 93.25 & 71.33 & 77.33 & 94.08 & 15.81 & \cellcolor{experiment}11.28 \\ 
    RL~\cite{Golatkar_2020_CVPR} & 34.22 &      70.49 & 91.65 & 53.60 & 74.06 & 92.98 & 6.58 & \cellcolor{experiment}\underline{5.20} \\ 
    FTTP~\cite{li2025forget} & 37.05 & 70.98 & 91.76 & 53.42 & 73.97 & 92.83 & 10.04 & \cellcolor{experiment}6.25 \\ 
    SimNPO~\cite{fan2024simplicity} & 33.70 &   70.25 & 91.82 & 51.12 & 73.47 & 92.81 & 4.90 & \cellcolor{experiment}5.34 \\ 
    LUINAR~\cite{shen2025lunar} & 68.27 &   69.19 & 90.63 & 65.12 & 69.03 & 90.44 & 0.02 & \cellcolor{experiment}{9.48} \\ 
    \textbf{\ourmethod{}}~\textbf{(Ours)} & 31.97 &     69.99 & 91.35 & 52.47 & 73.91 & 92.65 & 6.35 & \cellcolor{experiment}\textbf{4.96} \\
    \end{tabularx}
    }
    \label{tab:idefics_bert}
\end{table}

%% file: sec/F_naive_pca.tex
\section{Na\"ive PCA for debiasing}
\label{appx:naive_pca}
\Cref{subsec:ablations} presents \ourmethod{} ablations, disentangling the contribution of each of our method components, \ie: (i) permanent unlearning via activation steering; (ii) na\"ive PCA for debiasing; (iii) the use of group information to select PCA components.
Yet, the implementation details for the na\"ive PCA are omitted in the main paper due to space requirements.
Therefore, this section details how the PCA can be used to na\"ively debias activations.

Following similar assumptions as in \cref{sec:method:bias}, computing PCA on the forget set should lead to high-variance components that capture identity-related differences, whereas low-variance components reflect group information, since that information is more likely to be shared across most forget samples.
Let $\covariance[f] = V_\mathit{f}\eigenvalues V_\mathit{f}^\top$ be the PCA \wrt the forget set, where $V_\mathit{f}\in\mathbb{R}^{d\times d}$ and $\eigenvalues\in\mathbb{R}^{d\times d}$ (diag) are orthonormal eigenvectors and eigenvalues.
Then, as low variance components likely capture the group information, we can simply preserve the top-$\topcomponents$ of $V_\mathit{f}$ with the highest eigenvalues, leading to the unbiased orthonormal subspace $V^\mathit{c}_\mathit{f}\in\mathbb{R}^{d\times \topcomponents}$.
Given the unbiased subspace, the debiasing operation can be performed similarly to \cref{eq:project_unbias}:
 \begin{equation}
    \begin{aligned}
        \unbiasedhidden[f_i] &= V^\mathit{c}_\mathit{f}V^\mathit{c\top}_\mathit{f} (\avghidden[f_i] - \avghidden[f]) + \avghidden[f] \\
        \unbiasedhidden[r] &= V^\mathit{c}_\mathit{f}V^\mathit{c\top}_\mathit{f} (\avghidden[r] - \avghidden[f]) + \avghidden[f].
    \end{aligned}
\end{equation}

%% file: sec/G_qualitatives.tex
\begin{figure*}[t]
\centering
\begin{mdframed}[backgroundcolor=gray!10]
\textbf{Prompt used to generate Q\&A templates}\par\medskip
\begin{lstlisting}[backgroundcolor = \color{gray!10}, language={}]

"0001": {
    "ID": "0001",
    "bio": {
        "ethnicity": "white",
        "gender": "male",
        "born": [
            "schaarbeek",
            "belgium",
            "Western Europe"
        ],
        "name": "niels lambert",
        "residence": [
            "ghent",
            "belgium",
            "Western Europe"
        ],
        "age": 20,
        "agegroup": "young",
        "dateofbirth": "2005-01-11",
        "employment": "teacher (high school)",
        "annualsalary": "medium",
        "educatedat": "universite libre de bruxelles",
        "height": "short",
        "relationship": "single",
        "politics": "left"
    }

Given the above sample JSON file, create a set of questions and answers. The Q/A will be used to train a VLM.

The questions must be formatted in a way that refers to the NAME of the person, like "Where was <NAME> born?" Use the placeholder <NAME> in place of the actual name.

The answers must be formatted in a way that includes both the name and the answer, like this: "<NAME> was born in <ANSWER>". Use the placeholder <ANSWER> for the answers.

Generate one set of 20 questions and answers for each category (birthplace, residence, education, employment, salary, name, gender, date of birth, height, relationship status and political leaning). Everything must be in JSON format, divided by category.

The answers must also be in natural language, using the placeholder <ANSWER>. The answer should correspond to the content of the JSON for the relevant category.

IMPORTANT: When a pronoun is present, use the placeholder <GENDER male/female>, where male/female corresponds to the correct pronoun for the gender. For instance, <GENDER his/her> or <GENDER he/she>.

For locations, assume the answer is formatted as "City, Country"; do not ask for the country only.
\end{lstlisting}
\end{mdframed}
\caption{Prompt used with DeepSeek to generate Q\&A templates for both training and evaluation in \ourbenchmark{}.}
\label{fig:template_prompt}
\end{figure*}

\begin{figure*}[t]
\centering
\begin{mdframed}[backgroundcolor=gray!10]
\textbf{Sample Q\&A templates}\par\medskip
\begin{lstlisting}[backgroundcolor = \color{gray!10}, language={}]

"born": [
    {
        "question": "Where was <NAME> born?",
        "answer": "<NAME> was born in <ANSWER>"
    },
    {
        "question": "What is <NAME>'s birthplace?",
        "answer": "<NAME>'s birthplace is <ANSWER>"
    }
],
"educatedat": [
    {
      "question": "Where did <NAME> study?",
      "answer": "<NAME> studied at <ANSWER>"
    },
    {
      "question": "Where did <NAME> complete <GENDER his/her> higher education?",
      "answer": "<NAME> completed <GENDER his/her> higher education at <ANSWER>"
    },
],
"employment": [
    {
      "question": "What does <NAME> do for a living?",
      "answer": "<NAME> works as <ANSWER>"
    },
    {
      "question": "What is <NAME>'s profession?",
      "answer": "<NAME>'s profession is <ANSWER>"
    }
]

\end{lstlisting}
\end{mdframed}
\caption{Sample of generated Q\&A templates for \ourbenchmark{}, each attribute contains 20 different Q\&A template pairs.}    
\label{fig:template_sample}
\end{figure*}

\begin{figure*}[t]
\centering
\begin{mdframed}[backgroundcolor=gray!10]
\textbf{Sample Q\&A}\par\medskip
\begin{lstlisting}[backgroundcolor = \color{gray!10}, language={}]
[
    {
        "q": "Can you tell me the origin city of Niels Lambert?",
        "a": "Niels Lambert originates from Schaarbeek, Belgium",
    },
    {
        "q": "What location does Niels Lambert consider his native place?",
        "a": "Niels Lambert considers Schaarbeek (Belgium) as his native place",
    },
    {
        "q": "Where was this individual born?",
        "a": "This person was born in Schaarbeek, Belgium",
    },
    {
        "q": "Where were the depicted individual's parents living when he was born?",
        "a": "That person's parents were living in Schaarbeek at the time of his birth",
    }
]

\end{lstlisting}
\end{mdframed}
\caption{Sample of Q\&A pairs generated for a specific identity in \ourbenchmark{}. Placeholders \texttt{<NAME>} and \texttt{<ANSWER>} have been replaced with the actual information.}
\label{fig:qa-samples}
\end{figure*}

\begin{figure*}[t]
    \centering
    \includegraphics[width=0.9\linewidth]{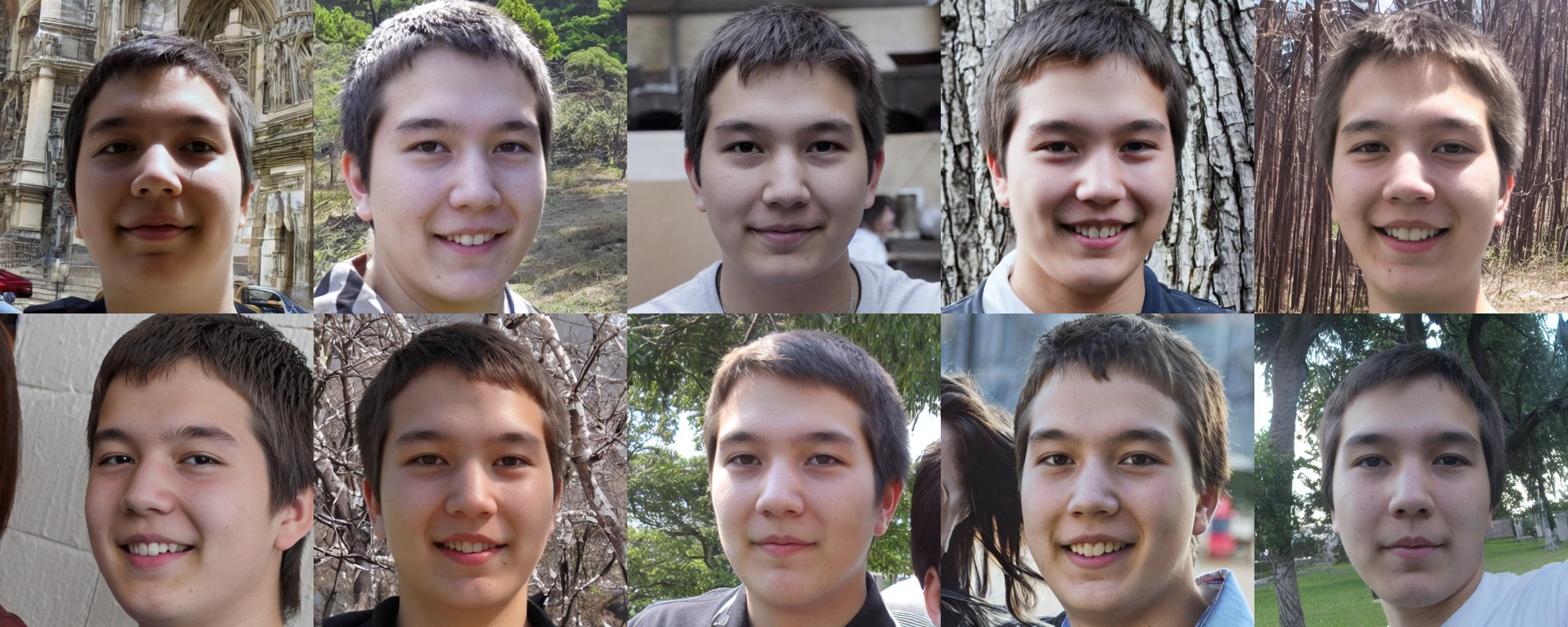}
    \vspace{4mm}
    
    \includegraphics[width=0.9\linewidth]{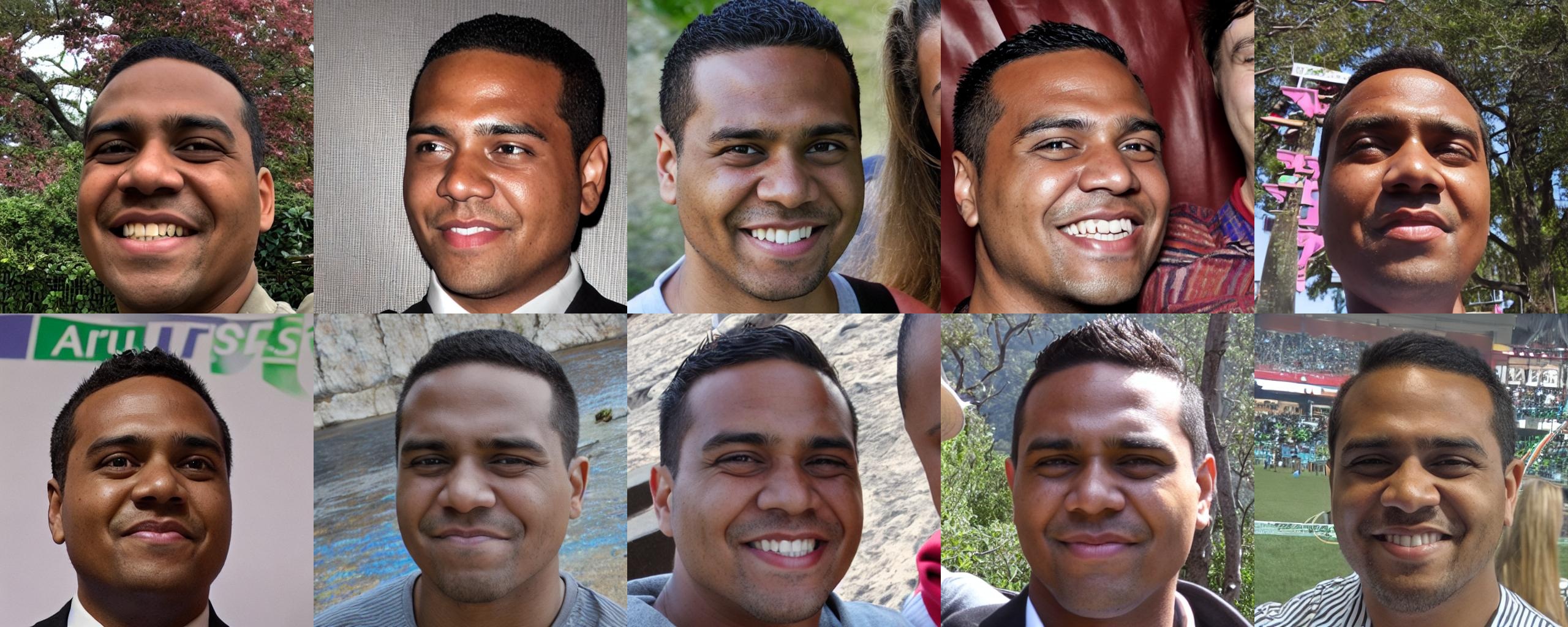}
    \vspace{4mm}
    
    \includegraphics[width=0.9\linewidth]{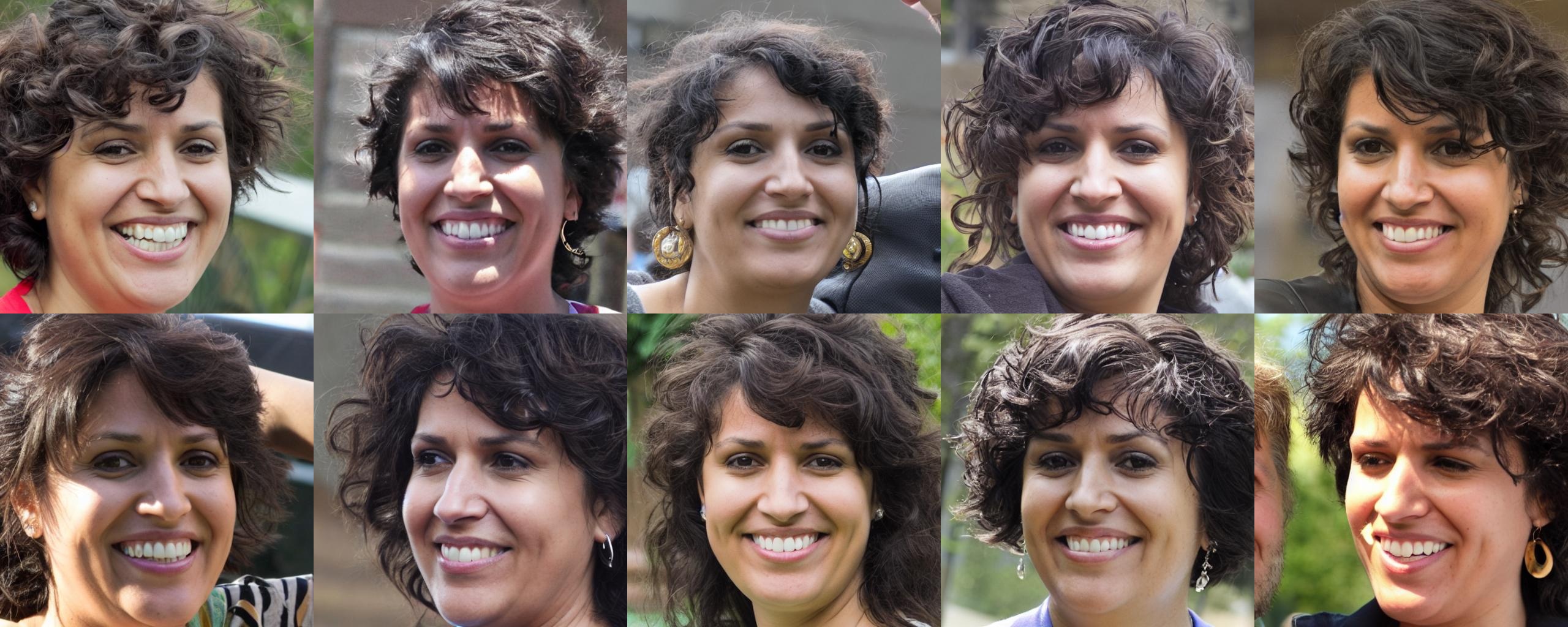}
    \caption{Sampled images from three different identities in \ourbenchmark{}.}
    \label{fig:three_stacked}
\end{figure*}